\documentclass[10pt,twocolumn,letterpaper]{article}

\usepackage[pagenumbers]{cvpr} 

\renewcommand{\paragraph}[1]{\vspace{.5em}\noindent\textbf{#1}}

\definecolor{cvprblue}{rgb}{0.21,0.49,0.74}
\usepackage[pagebackref,breaklinks,colorlinks,allcolors=cvprblue]{hyperref}

\def\paperID{16202} 
\def\confName{CVPR}
\def\confYear{2026}

\title{FarSLIP: Discovering Effective CLIP Adaptation for Fine-Grained Remote Sensing Understanding}

\author{
Zhenshi Li$^{1,2}$\hspace{0.5cm} 
Weikang Yu$^{3}$\hspace{0.5cm} 
Dilxat Muhtar$^{1}$\hspace{0.5cm} 
Xueliang Zhang$^{1,}$\thanks{Corresponding author.}\hspace{0.5cm} 
Pengfeng Xiao$^{1}$,\\[0.25em]
Pedram Ghamisi$^{3}$\hspace{0.5cm} 
Xiao Xiang Zhu$^{2,*}$\\[0.6em]
{\fontsize{10pt}{11pt}\selectfont
$^{1}$Nanjing University \quad
$^{2}$Technical University of Munich \quad
$^{3}$Helmholtz-Zentrum Dresden-Rossendorf
}\\
{\tt\small Lzhenshi@outlook.com, zxl@nju.edu.cn, xiaoxiang.zhu@tum.de} 
}

\usepackage{multirow}
\newcommand{\dn}{MGRS-200k}
\newcommand{\mn}{FarSLIP}
\usepackage{pifont}
\newcommand{\cmark}{\ding{51}} 
\newcommand{\xmark}{\ding{55}}

\usepackage{amsmath}
\usepackage{tcolorbox}
\usepackage{xcolor}

\newcommand{\Lglo}{$\mathcal{L}_{\mathrm{glo}}$}
\newcommand{\Lloc}{$\mathcal{L}_{\mathrm{loc}}$}
\newcommand{\Ldis}{$\mathcal{L}_{\mathrm{dis}}$}

\begin{document}
\maketitle
\begin{abstract}
As CLIP’s global alignment limits its ability to capture fine-grained details, recent efforts have focused on enhancing its region-text alignment.
However, current remote sensing (RS)-specific CLIP variants still inherit this limited spatial awareness.
We identify two key limitations behind this: (1) current RS image-text datasets generate global captions from object-level labels, leaving the original object-level supervision underutilized; (2) despite the success of region-text alignment methods in general domain, 
their direct application to RS data often leads to performance degradation.
To address these, we construct the first multi-granularity RS image-text dataset, \dn,
featuring rich object-level textual supervision for RS region-category alignment.
We further investigate existing fine-grained CLIP tuning strategies and find that current explicit region-text alignment methods, whether in a direct or indirect way, underperform due to severe degradation of CLIP’s semantic coherence.
Building on these, we propose \mn, a \textbf{F}ine-grained \textbf{A}ligned \textbf{RS} \textbf{L}anguage-\textbf{I}mage \textbf{P}retraining framework. 
Rather than the commonly used patch-to-CLS self-distillation, FarSLIP employs patch-to-patch distillation to align local and global visual cues, which improves feature discriminability while preserving semantic coherence.
Additionally, to effectively utilize region-text supervision, it employs simple CLS token-based region-category alignment rather than explicit patch-level alignment, further enhancing spatial awareness.
\mn~features improved fine-grained vision-language alignment in RS domain and sets a new state of the art not only on RS open-vocabulary semantic segmentation, but also on image-level tasks such as zero-shot classification and image-text retrieval.
Our dataset, code, and models are available at \href{https://github.com/NJU-LHRS/FarSLIP}{https://github.com/NJU-LHRS/FarSLIP}.

\end{abstract}    
\section{Introduction}
\label{sec:intro}

Vision-language foundational models (VLFMs), such as CLIP \cite{radford2021clip}, have exhibited remarkable generalizability across diverse downstream tasks.
Thanks to the rich set of real-world concepts encoded in VLFMs, they can be effectively transferred to other domains, achieving significantly higher accuracy than training from scratch \cite{mukhoti2023fine,kumar2022fine}. 
In the remote sensing (RS) community, promising in-domain performance is obtained by building RS-specific image-caption datasets and fully fine-tuning CLIP \cite{liu2024remoteclip, wang2024skyscript}, a strategy shown to be effective for CLIP’s domain adaptation \cite{goyal2023finetune}.

Recent studies have shown that CLIP inherently exhibit superior localization capabilities, enabling training-free open-vocabulary semantic segmentation (OVSS) \cite{zhou2022maskclip, wang2024sclip,lan2024clearclip,lan2024proxyclip}. This intrinsic capability also holds great potential for supporting domain-specific dense prediction tasks, such as land cover mapping in RS imagery \cite{li2025segearth} with more flexible and adaptable category systems. 
However, since CLIP is trained by aligning global image and text features, it still exhibits limited capability in achieving high-precision spatial recognition \cite{wu2024clipself,wysoczanska2024clipDINOise,barsellotti2025talkingtoDINO}.
Therefore, recent studies have further aimed to enhance its inherent fine-grained associations by explicitly aligning region-level features with language. 
Approaches such as weakly supervised alignment \cite{mukhoti2023pacl, sparc}, large-scale object-text dataset construction \cite{zhong2022regionclip, chen2024cloc}, and self-distillation \cite{wu2024clipself, yeo2025atas} have been proposed to further improve the performance on fine-grained tasks.

\begin{figure}[ht]
  \centering
  \includegraphics[width=\linewidth]{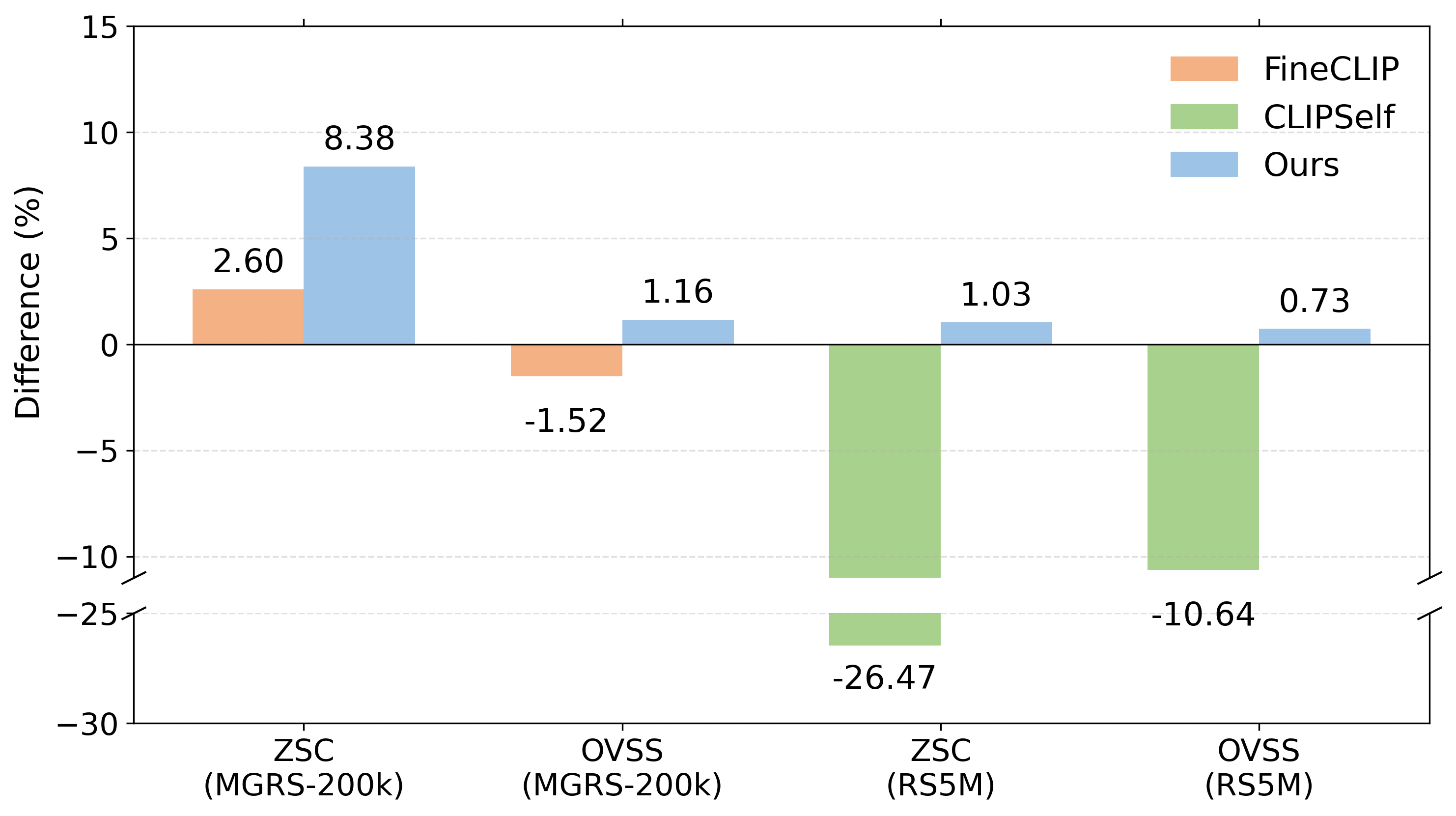}
  \caption{Accuracy differences relative to CLIP for models fine-tuned with SOTA methods, \eg, FineCLIP \cite{jing2024fineclip} and CLIPSelf \cite{wu2024clipself}, on various RS datasets (shown in brackets). These models show only marginal improvement or even performance drops in most cases, indicating limited effectiveness for RS-domain application.}
  \label{fig1}
\end{figure}

However, despite being specifically optimized for fine-grained recognition, these general-domain models still perform poorly on OVSS in RS images, even worse than the RS variants trained with simple vision-language contrastive learning (CL), as shown in \cref{tab:final ovss}, highlighting the importance of RS-specific fine-tuning.
Furthermore, we find that these strategies are suboptimal when directly applied to RS domain.
As shown in \cref{fig1}, directly applying state-of-the-art (SOTA) fine-grained CLIP adaptation methods to RS datasets even fails to surpass the original CLIP baseline.

To better equip VLFMs for fine-grained RS interpretation, we consider current limitations from two aspects: data and methodology.
On the data side, due to the inaccessibility of large-scale web-crawled image-text data for RS, existing RS datasets are typically constructed by generating global-level captions based on object-level information, sourced from object detection or segmentation datasets \cite{liu2024remoteclip, xiong2025geolangbind}, or from crowd-sourced geospatial database \cite{wang2024skyscript,muhtar2024lhrs}. 
We argue this paradigm underutilizes the rich regional cues in the original data as a source of fine-grained supervision. 
Hence, to advance fine-grained CLIP tuning in RS, we propose \dn, the first multi-granularity RS image-text dataset. 
Beyond global captions, as commonly seen in prior works, \dn~features over 1M object-category annotations, supporting explicit region-level vision-language alignment for RS-domain CLIP tuning.

On the methodology side, motivated by the side effects of existing fine-grained CLIP tuning methods on RS (\cref{fig1}), we conduct a comprehensive analysis to understand their underlying causes.
We find that current explicit region-text alignment approaches, whether direct patch-to-text alignment \cite{jing2024fineclip,xie2025FG-CLIP} or indirect alignment via patch-to-CLS self-distillation \cite{wu2024clipself,yeo2025atas,jing2024fineclip}, are in fact detrimental to CLIP tuning, as they substantially disrupt the model’s semantic coherence, a property shown to be crucial for fine-grained understanding \cite{oquab2024dinov,wysoczanska2024clipDINOise,lan2024proxyclip,qiu2025scrla}.
In order to overcome this, we identified that:
(1) For direct region-category alignment via CL, tuning CLIP by aligning text with the CLS token rather than patch features \cite{jing2024fineclip,xie2025FG-CLIP} not only improves feature discriminability and region-text alignment but also enhances semantic coherence;
(2) For indirect region-text alignment via local-global self-distillation, we propose patch-to-patch distillation instead of the commonly used patch-to-CLS distillation \cite{wu2024clipself,yeo2025atas,jing2024fineclip}, preserving semantic coherence while enhancing feature discriminability.

Following the above insights, we develop an RS-specific VLFM called \mn. By employing the above-mentioned appropriate tuning strategies with the proposed multi-granularity RS image-text dataset, we enhance the CLIP’s understanding capabilities at both global and local levels for RS tasks. \mn~achieves SOTA performance across both fine-grained tasks (\eg, OVSS) and global-level tasks such as zero-shot classification (ZSC) and cross-modal retrieval.

In summary, our contributions are threefold:

\begin{itemize}
\item We construct \dn, the first multi-granularity image-text dataset for RS, establishing a foundation for fine-grained RS-specific VLFM training.
\item We conduct a thorough analysis of fine-grained CLIP tuning for RS, offering transferable insights for domain-specific fine-grained adaptation of CLIP.
\item We propose \mn, an RS-specialized VLFM that achieves SOTA performance on both global- and local-level RS multimodal tasks.

\end{itemize}

\section{Related Work}
\label{sec:related work}

\subsection{Fine-grained Vision-Language Alignment}
To enhance VLFMs’ fine-grained vision-language alignment, recent studies have explored self-supervised \cite{maninis2024tips, naeem2024silc, tschannen2025siglip2, kim2025cosmos} or weakly supervised objectives \cite{sparc, mukhoti2023pacl,asokan2025finelip}.
Given the high cost of training from scratch, most existing works focus on leveraging the pretrained capabilities of CLIP.
One line of work explores self-distillation to exploit the frozen CLIP model’s own knowledge \cite{wu2024clipself,qiu2025scrla,yeo2025atas,jing2024fineclip}, while another constructs large-scale region-text datasets to improve local-level alignment in feature space \cite{zhong2022regionclip,chen2024cloc,jing2024fineclip,xie2025FG-CLIP}.
However, these models lag behind RS CLIP variants without fine-grained training (\cref{tab:final ovss}), and even when trained on RS data, these methods achieve only limited improvements or introduce side effects (\cref{fig1}).
How to effectively adapt CLIP for fine-grained RS interpretation remains unresolved.

\subsection{Vision-Language Foundation Models for RS}
Recent studies have built RS image-caption datasets to adapt CLIP with domain-specific knowledge, producing RS-specialized VLFMs \cite{liu2024remoteclip, wang2024skyscript, zhang2024rs5m}.
However, their focus on global-level CL inherits CLIP's limitations in fine-grained visual understanding. 
Moreover, due to the unavailability of large-scale RS image-text data on the web, current RS datasets commonly synthesize global captions by inputting region-level annotations, sourced from object detection, semantic segmentation datasets \cite{liu2024remoteclip, xiong2025geolangbind} or geospatial information like OpenStreetMap \cite{wang2024skyscript, muhtar2024lhrs, li2025lhrs}, into large language models (LLMs) or multimodal LLMs.
We argue that these approaches significantly underutilize the rich spatial and semantic object-level annotations already available in their source data. These annotations, if explicitly preserved, could serve as powerful supervision signals for enhancing fine-grained vision-language alignment, a potential that remains unexplored in the RS domain.

\section{Multi-Granularity RS Image-Text Dataset}

Unlike prior works that generate only global captions while discarding local object information \cite{liu2024remoteclip, xiong2025geolangbind,wang2024skyscript, muhtar2024lhrs, li2025lhrs}, we explicitly preserve spatial and semantic details of individual geographical objects. Based on this, we propose \dn, the first RS image-text dataset designed for multi-granularity vision-language alignment.

\dn~is constructed from 16 RS object detection datasets, sourced either from public data or generated using multiple foundation models on RS images. 
Conditioned on the RS images along with their corresponding object locations and categories, we employ Intern-VL3 \cite{zhu2025internvl3} to generate/recaption detailed
captions for each image.
Inspired by recent studies \cite{kim2025cosmos, choi2025goal, asokan2025finelip, zheng2024dreamlip, xiao2025flair} suggesting that longer and more detailed descriptions can benefit fine-grained alignment, we generate two types of captions for each image: a detailed but short caption and a long and more comprehensive caption.
As a result, \dn~contains about 200k RS images, each paired with both a short and a long global-level caption, as well as multiple object bbox-category pairs, in total, over one million instances.

We assess hallucinations within the dataset and demonstrate its reliability. We further show that our generated captions achieve significant improvements compared to existing RS datasets. Please refer to Appendix for details on data sources, caption generation, illustrative examples, data quality assessment, and additional ablation studies.

\section{Fine-grained Image-Text Alignment in RS}
\label{sec:analysis}

\begin{figure}[ht]
  \centering
  \includegraphics[width=\linewidth]{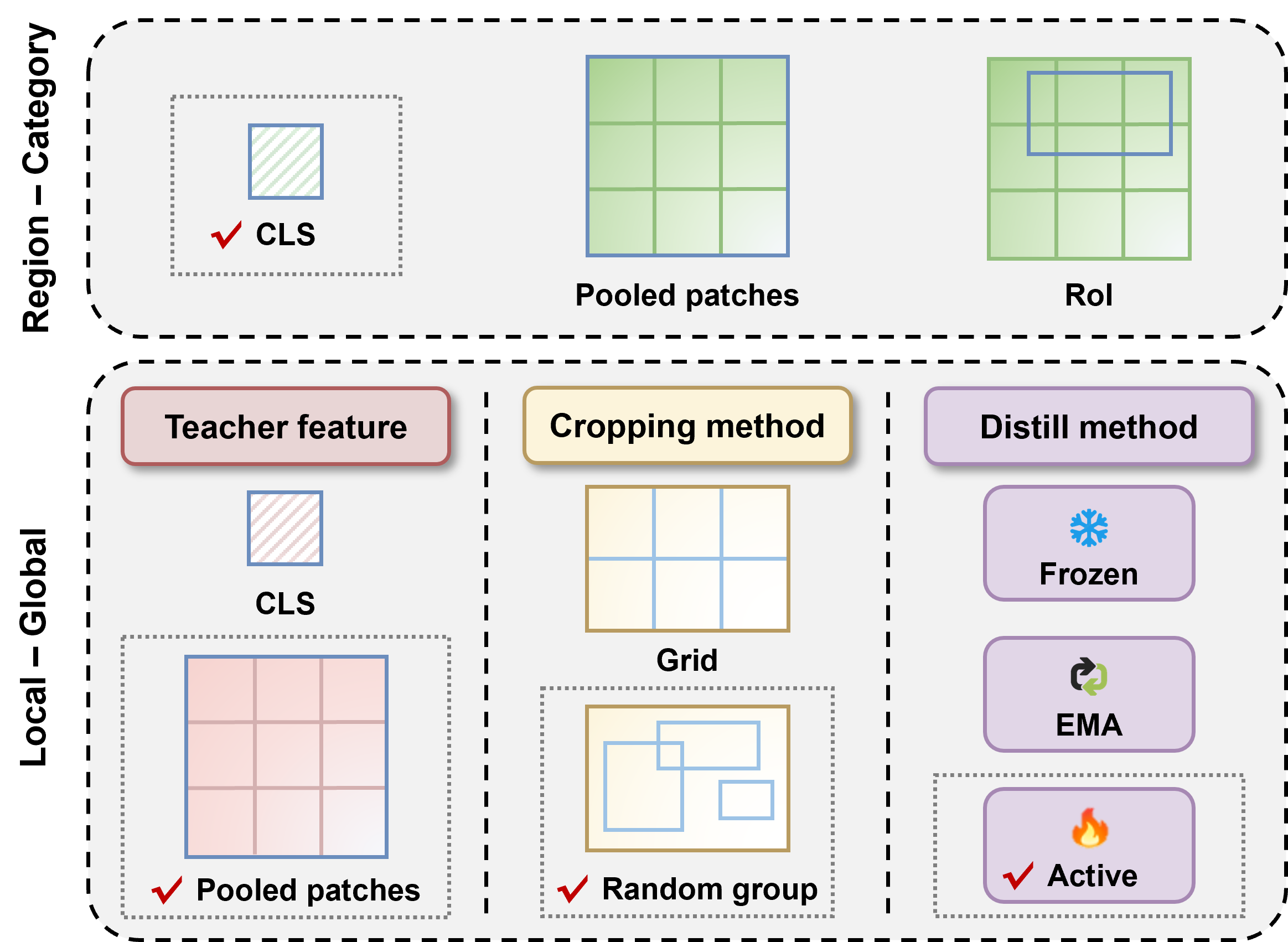}
  \caption{Overview of the analysis procedure. The item marked with a checkmark indicates the most effective method identified for RS-specific CLIP tuning.}
  \label{fig:analysis}
\end{figure}

Although fine-grained tuning methods for CLIP have proven effective in general domains,
they often yield limited or negative transfer when applied to current RS data, as shown in \cref{fig1}. 
This motivates a systematic investigation into effective training strategies tailored for RS-specific fine-grained vision-language alignment.
To this end, we conduct an in-depth analysis on two key aspects: \textit{region-category alignment} and \textit{local-global alignment}. The former directly aligns patch embeddings with text via region-text CL \cite{jing2024fineclip, xie2025FG-CLIP}, while the latter achieves indirect region-text alignment via local-global self-distillation \cite{wu2024clipself,jing2024fineclip,yeo2025atas}.
In this section, we first present the overall architecture of our model, followed by the experimental setup, and finally delve into the two analytical perspectives outlined above.
The overall analytical framework is illustrated in \cref{fig:analysis}.

\subsection{Model Architecture}

\begin{figure}[!t]
  \centering
  \includegraphics[width=\linewidth]{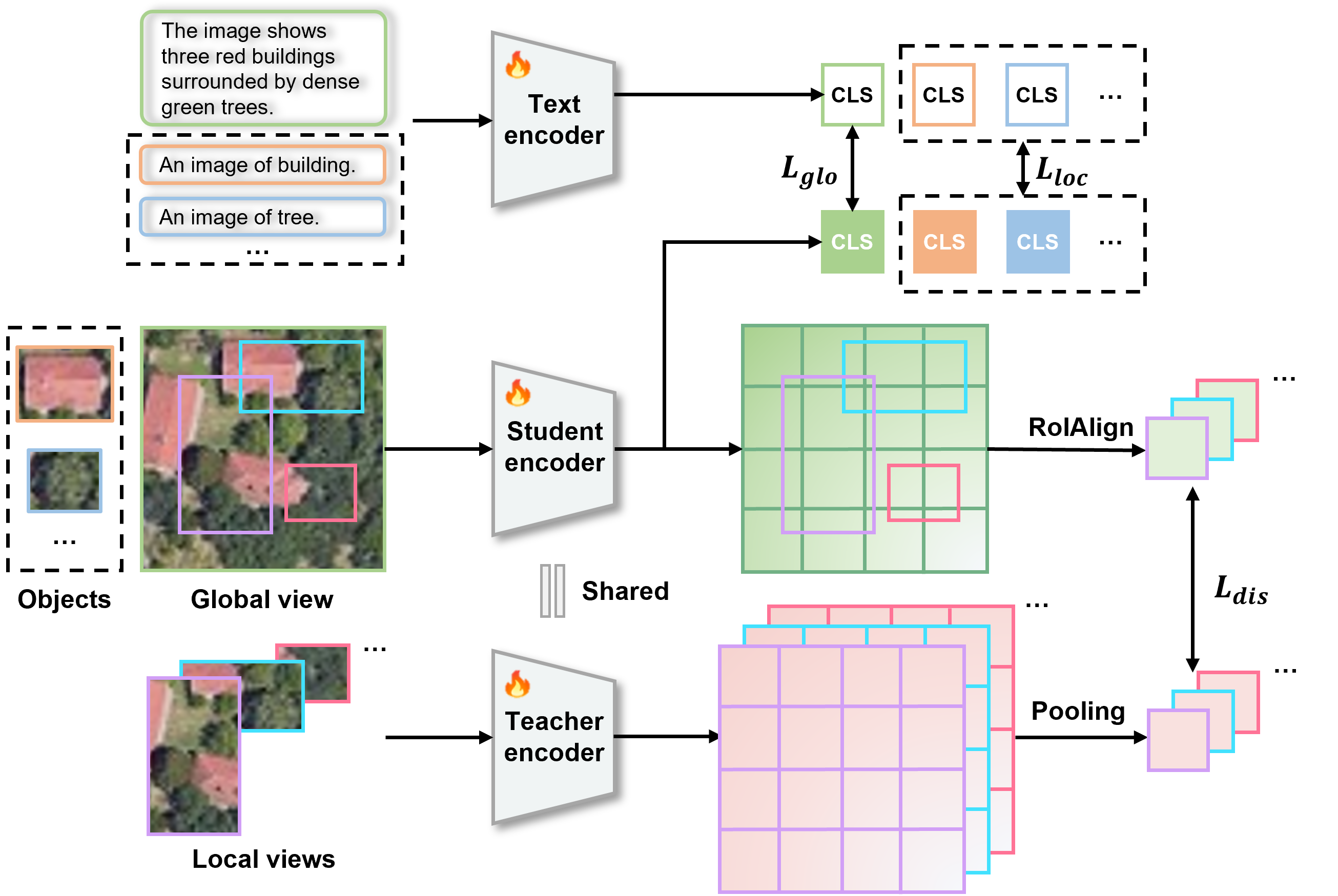}
  \caption{Architecture of \mn. Stage one is trained on image-caption data with \Lglo~and \Ldis, while stage two leverages image-caption and object-category pairs with \Lglo~and \Lloc.}
  \label{fig:architecture}
\end{figure}

Our model is built upon the CLIP architecture \cite{radford2021clip} for broad pre-trained model compatibility.
Self-distillation that aligns global and local visual representations \cite{wu2024clipself, yeo2025atas} is employed to enhance fine-grained vision-language alignment.
Specifically, we employ a teacher-student framework with two vision encoders sharing the same CLIP vision backbone. 
For image-text alignment, the student extracts image features to perform CL with text embeddings. For self-distillation, the input image is cropped into local views for the teacher, while the student processes the original full image. The features from the two encoders are then aligned through a local-global distillation loss. During inference, only the student encoder is used to extract visual features. An overview of the model is shown in \cref{fig:architecture}, 
with its components described in detail in \cref{sec:rca} and \cref{sec:lga}.

\subsection{Experimental Setup}

\paragraph{Model training.} 
Our training data consists of the RS5M dataset \cite{zhang2024rs5m}, which provides image-caption pairs, and our proposed \dn~dataset, which contains both image-caption and region-category pairs. We use the short captions in \dn~for image-text alignment.

For image-caption pairs, we follow CLIP and employ the standard InfoNCE loss to achieve global-level alignment between image and text embeddings. 
Specifically, the global loss for a batch of $N$ image-text pairs is defined as:

\begin{equation}
\mathcal{L}_{\text{glo}} = \frac{1}{2} \left( \mathcal{L}_{I \rightarrow T} + \mathcal{L}_{T \rightarrow I} \right)
\end{equation}
\begin{equation}
\mathcal{L}_{I \rightarrow T} = -\frac{1}{N} \sum_{i=1}^{N} \log \frac{\exp(S(V_i, T_i)/\tau)}{\sum_{j=1}^{N} \exp(S(V_i, T_j)/\tau)}
\end{equation}
where \( V_i \) and \( T_i \) denote the image and text \texttt{[CLS]} embeddings of the \( i \)-th matched pair in a batch, respectively; the function \( S(\cdot, \cdot) \) computes the cosine similarity between the embeddings; and \( \tau \) is a learnable temperature.

For region-category pairs in the \dn~dataset, the category labels are first converted into textual descriptions using a predefined template, such as ``a satellite image of \{\textit{category}\}". These descriptions are then used in the CL loss. However, since the same category often appears within a single image or across multiple samples, treating other samples with the same category as negative examples is semantically inappropriate. To address this, we adopt multi-positive CL (MPCL) \cite{khosla2020SupCon,tian2023stablerep}, where all region-category pairs sharing the same category are treated as positive matches. This allows the model to better capture the semantic consistency within each category. 
Specifically, the MPCL for a batch of \( M \) region-category pairs is defined as:

\begin{equation}
\mathcal{L}_{\text{loc}} = \frac{1}{2} \left( \mathcal{L}_{R \rightarrow C} + \mathcal{L}_{C \rightarrow R} \right)
\end{equation}
{
\fontsize{9pt}{\baselineskip}\selectfont
\begin{equation}
\mathcal{L}_{R \rightarrow C} = \frac{1}{M} \sum_{i=1}^{M} -\frac{1}{|P(i)|} \sum_{j \in P(i)} \log 
\frac{\exp\bigl(S(V_i^r, T_j^c)/\tau\bigr)}{\sum_{k=1}^{M} \exp\bigl(S(V_i^r, T_k^c)/\tau\bigr)}
\end{equation}
}
\noindent where \( V_i^r \) is the visual embedding of the \( i \)-th region; \( T_j^c \) is the textual \texttt{[CLS]} embedding of the \( j \)-th category; and the set \( P(i) \) contains indices of positive samples sharing the same category as region \( i \).
The specific choice of visual embedding will be discussed in \cref{sec:rca}.

For the self-distillation process, the full image is passed through the student encoder. 
Given the bounding boxes of image crops, we apply RoIAlign \cite{he2017maskrcnn} on the patch tokens to extract corresponding region-specific features, which we refer to as RoI features $\mathbf{p}_i^{\mathrm{roi}}$. 
The local view, which is a cropped region from the original image, is fed into the teacher encoder to extract the corresponding local features $\mathbf{p}_i^{\mathrm{local}}$. Then, the local features and the RoI features are used to calculate the distillation loss as follows.

\begin{equation}
\mathcal{L}_{\text{dis}} = \frac{1}{m} \sum_{i=1}^m \left( 1 - \frac{\mathbf{p}_i^{\mathrm{roi}} \cdot \mathbf{p}_i^{\mathrm{local}}}{|\mathbf{p}_i^{\mathrm{roi}}| \, |\mathbf{p}_i^{\mathrm{local}}|} \right)
\end{equation}

Our image and text encoders are initialized with the pretrained weights from OpenAI CLIP \cite{radford2021clip}. 
More details about model training are provided in Appendix.
Notably, we observe that when fine-tuned on RS image-text datasets, CLIP models based on ViT-L/14 consistently underperform those based on ViT-B/16. Therefore, we focus our experiments on ViT-B/16 and ViT-B/32 throughout this study.

\paragraph{Task, benchmark, and evaluation.}
Following prior works \cite{liu2024remoteclip,wang2024skyscript}, we evaluate global-level image-text alignment on eight RS scene classification datasets \cite{wang2024skyscript,xia2017aid,helber2019eurosat,christie2018fmow,long2021millionaid,zhou2018patternnet,cheng2017RESISC45,li2020rsicb}, and three cross-modal retrieval datasets \cite{lu2017rsicd,yuan2022rsitmd,yang2010ucm}.
For local-level alignment evaluation, we employ eight RS semantic segmentation datasets \cite{xia2023openearthmap,wang2021loveda,waqas2019isaid,Potsdam,Vaihingen,lyu2020uavid,chen2018udd,cai2025vdd}.
The evaluation metrics include Top-1 accuracy for ZSC, mean recall at R@1, R@5, and R@10 for cross-modal retrieval \cite{liu2024remoteclip}, and mIoU for OVSS.
The implementation of OVSS follows recent training-free methods \cite{zhou2022maskclip, li2025segearth}, with the specific approach provided in Appendix.
All reported results are averaged across the corresponding datasets.

\subsection{Region-Category Alignment}
\label{sec:rca}

\paragraph{Vision feature for CL.}
CLIP performs global alignment between image and text features, which introduces a gap when applied to dense prediction tasks such as OVSS, where region-level features are required. 
Recent works have proposed explicit region-text alignment to address this issue \cite{jing2024fineclip, xie2025FG-CLIP}, by aligning patch-level visual features (\( \mathbf{p}^{\mathrm{roi}} \)) with textual CLS tokens via CL.
But does direct region-text alignment remain effective when fine-tuning CLIP for RS?

Leveraging the region-category pairs in \dn, we compare three types of visual embeddings for region-category alignment using \Lloc: (1) the CLS token of the region image; (2) the average-pooled patch tokens of the region image; and (3) the RoI feature \( \mathbf{p}^{\mathrm{roi}} \), extracted from the global image patch tokens using RoIAlign over the region extent, following the method in FineCLIP~\cite{jing2024fineclip}.

Our findings differ significantly from those reported in FineCLIP, which benefits from more explicit region-text alignment. As shown in \cref{tab:local-feature-vit}, RoI feature-based alignment leads to a notable performance drop compared to using CLS token, even underperforming the baseline.
This trend consistently appears in both ZSC and OVSS tasks, even though OVSS is expected to benefit more from direct region-level supervision.
A similar pattern is observed with pooled patch tokens: despite incorporating spatial cues, they still underperform on OVSS compared to CLS tokens.

To better understand this phenomenon, we investigate the models’ inherent capabilities. We argue that a VLFM with strong fine-grained understanding abilities should produce visual features with three properties: (1) clear feature boundaries, \ie, high inter-class separation and low intra-class variance; (2) accurate region-language alignment, \ie, accurately assigning region features to their corresponding textual categories; and (3) high semantic coherence, \ie, pixels with similar content should have similar features.
To quantify these abilities, we use the Davies-Bouldin Index (DBI) \cite{davies2009DBI} to indicate feature discriminability, compute top-1 accuracy (Acc@1) of instance features against text queries to represent region-text alignment, and calculate the mean average precision (mAP) of feature similarity across pixels to reflect semantic coherence. Details of the computation of these metrics are provided in the appendix.

We compare the analysis metrics of region-text alignment using RoI features and CLS tokens, as shown in \cref{tab:deepin}. Thanks to RS-specific fine-tuning, both approaches improve the model’s feature discriminability and region-language alignment, as reflected by the increased DBI and Acc@1. 
The RoI-based method performs slightly better, due to its more explicit instance-level training. However, this approach significantly reduces the model’s semantic coherence, which is critical for fine-grained understanding \cite{oquab2024dinov,wysoczanska2024clipDINOise,lan2024proxyclip}, thereby leading to worse OVSS performance.
In comparison, CLS token-based training not only preserves the improved feature discriminability and region-text alignment but also enhances semantic coherence, which is one of the inherent properties of CLIP \cite{zhou2022maskclip,lan2024proxyclip}. 
Additional qualitative analyses can be found in the appendix.

\begin{tcolorbox}[
    colback={rgb,255:red,240;green,249;blue,253},   
    colframe={rgb,255:red,103;green,149;blue,206},  
    rounded corners,        
    boxrule=0.5mm,         
    left=2mm, right=2mm, top=0.5mm, bottom=0.5mm 
]
\textbf{Takeaway 1}: A more suitable approach is to preserve CLIP’s original visual-language association, \ie, aligning via the CLS token, while leveraging its inherent spatial awareness for RS OVSS \cite{zhou2022maskclip,wang2024sclip}.
\end{tcolorbox}

\begin{table}[ht]
\caption{Comparison of region visual feature types under \Lloc-only training on \dn~(baseline: CLIP).}
\centering
\begin{tabular}{l|cc|cc}
\toprule
\multirow{2}{*}{Region feat.} & \multicolumn{2}{c|}{ViT-B/16} & \multicolumn{2}{c}{ViT-B/32} \\
 & ZSC & OVSS & ZSC & OVSS \\
\midrule
Baseline & 50.60 & 34.52 & 43.10 & 29.32 \\ \midrule
RoI embedding    & 45.43 & 30.42 & 42.86 & 26.37 \\
Pooled patches & 48.12 & 28.89 & 45.44 & 26.69 \\
CLS token   & \textbf{58.69} & \textbf{35.42} & \textbf{57.02} & \textbf{29.38} \\
\bottomrule
\end{tabular}
\label{tab:local-feature-vit}
\end{table}

\begin{table}[htbp]
\setlength{\tabcolsep}{1mm}
\caption{Accuracy improvement percentages (unit: \%) over the baseline for different training methods. The first two rows show the relative changes achieved by fine-tuning with different region-category alignment methods compared to CLIP. The last two rows show the relative changes obtained by using different self-distillation methods versus no self-distillation.}
\centering
\begin{tabular}{llccc}
\toprule
Objective & Method & DBI & Acc@1 & mAP \\
\midrule
\multirow{2}{*}{Region-cat. align.} 
        & RoI embedding  & +21.8 & +1.1 & --1.5 \\
        & CLS token      & +18.3 & +0.5 & +1.2 \\
\midrule
\multirow{2}{*}{Self-distillation} 
        & RoI-to-CLS     & +0.9  & --3.9 & --1.2 \\
        & RoI-to-Pooled  & +6.1  & --0.6 & --0.2 \\
\bottomrule
\end{tabular}
\label{tab:deepin}
\end{table}


\paragraph{Effectiveness of region-category pairs.}
Unlike prior works that discard local information when generating global captions for RS images, we investigate the effect of retaining region-category pairs during dataset construction.
As \cref{tab:loss} shows, applying region-category CL significantly improves ZSC performance over using global image-caption CL. It also brings accuracy gains for OVSS on ViT-B/16, though a slight drop is observed on ViT-B/32, possibly due to small regions in the training dataset combined with coarse patch features. Nevertheless, combining \Lglo~and \Lloc~losses yields the best overall performance, highlighting the value of preserving region-category pairs when constructing RS image-text datasets. 
\begin{tcolorbox}[
    colback={rgb,255:red,240;green,249;blue,253},   
    colframe={rgb,255:red,103;green,149;blue,206},  
    rounded corners,        
    boxrule=0.5mm,         
    left=2mm, right=2mm, top=0.5mm, bottom=0.5mm 
]
\textbf{Takeaway 2}: Future RS image-text data generation should explicitly retain fine-grained supervision signals that previous RS datasets have underutilized.
\end{tcolorbox}

\begin{table}[t]
\caption{Comparison of different loss configurations on models fine-tuned with the \dn~dataset.
}
\centering
\begin{tabular}{cc|cc|cc}
\toprule
\multirow{2}{*}{\Lglo} & \multirow{2}{*}{\Lloc} & \multicolumn{2}{c|}{ViT-B/16} & \multicolumn{2}{c}{ViT-B/32} \\
       &        & ZSC & OVSS & ZSC & OVSS \\
\midrule
\cmark &        & 56.92 & 35.23 & 54.04 & \textbf{30.64} \\     
       & \cmark & 58.69 & 35.42 & 57.02 & 29.38 \\
\cmark & \cmark & \textbf{59.17} & \textbf{35.67} & \textbf{58.34} & 30.56 \\ 
\bottomrule
\end{tabular}
\label{tab:loss}
\end{table}

\subsection{Local-Global Alignment}
\label{sec:lga}

As self-distillation has been proven effective in indirectly enhancing region-text alignment through local-global distillation on top of CLIP, we conduct an in-depth analysis of its effectiveness for RS-specific tuning. 
We primarily use the RS5M dataset \cite{zhang2024rs5m} for the analysis, as its relatively large scale allows for more reliable conclusions. To analyze the effect of different data scales, we randomly sample two subsets: 0.5M and 2.5M image-text pairs. To balance data volume and experimental efficiency, experiments are conducted using the 2.5M subset unless otherwise specified.

\paragraph{What to distill?}
We investigate which teacher representations should be transferred to the student for effective self-distillation, focusing on selecting appropriate local representations and how to generate local views.

While prior studies demonstrate that self-distillation by aligning RoI features with cropped images’ CLS tokens greatly enhances local-level vision-language alignment \cite{wu2024clipself,jing2024fineclip,yeo2025atas}, our experiments reveal an unexpected inconsistency.
As shown in \cref{tab:distill feat}, CLS token-based self-distillation results in a noticeable performance drop in both ZSC and OVSS, which is surprising, particularly given the decline in OVSS accuracy, contradicting existing findings \cite{wu2024clipself,jing2024fineclip,yeo2025atas}. 

To further understand this phenomenon, we find that CLS-token-based self-distillation slightly improves feature discriminability (\cref{tab:deepin}), but largely degrades region-text alignment and semantic coherence.
This can be attributed to the nature of RS imagery: large spatial coverage, small targets, and highly heterogeneous scenes.
We observe that the CLS token often corresponds to the dominant regions in RS images (see appendix for visualizations).
In this case, aligning CLS tokens with RoI features leads the model to learn patch-level representations biased toward these dominant regions, while neglecting numerous non-dominant areas, thereby hindering fine-grained feature extraction.

To enable effective local-global self-distillation, we propose patch-to-patch distillation.
Specifically, as shown in \cref{fig:architecture}, we extract average-pooled patch embeddings from cropped images to perform patch-to-patch self-distillation.
This method has three advantages: (1) it aligns patch features with the content of their own crop views, thereby reducing background interference \cite{qiu2025scrla}; (2) it aligns patch features with higher-resolution representations extracted by the teacher model, improving precision \cite{fu2024featup,li2025segearth}; and (3) both views are derived from fine-grained patch features that preserve detailed scene information, avoiding information loss caused by the discriminative nature of the CLS token \cite{wu2024clipself,jing2024fineclip}.
\Cref{tab:distill feat} shows that this method is far more effective than CLS token-based distillation and improves results on both ZSC and OVSS tasks. It also leads to better feature discriminability while preserving semantic coherence, as evidenced in \cref{tab:deepin}.


Regarding the generation of local views, we compare two strategies: grid sampling \cite{wu2024clipself, yeo2025atas} and random cropping \cite{qiu2025scrla}. 
As shown in \cref{tab:distill feat}, 
random cropping is a better local view generation method in this context, demonstrated by higher OVSS accuracy and competitive ZSC performance.
\begin{tcolorbox}[
    colback={rgb,255:red,240;green,249;blue,253},   
    colframe={rgb,255:red,103;green,149;blue,206},  
    rounded corners,        
    boxrule=0.5mm,         
    left=2mm, right=2mm, top=0.5mm, bottom=0.5mm 
]
\textbf{Takeaway 3}: Patch-to-patch self-distillation is superior to CLS-to-patch, which tends to constrain patch features to dominant categories and lose visual details.
\end{tcolorbox}

\begin{table}[ht]
\caption{Comparison of distillation features and cropping methods. RC = random crop; CLS / pooled = CLS token / pooled patch token.}
\centering
\setlength{\tabcolsep}{1.2mm}
\begin{tabular}{l|cc|cc}
\toprule
\multirow{2}{*}{Distill feat. / Crop method} & \multicolumn{2}{c|}{ViT-B/16} & \multicolumn{2}{c}{ViT-B/32} \\
                              & ZSC & OVSS & ZSC & OVSS \\
\midrule
Without Distillation          & 64.13 & 36.34 & 62.22 & 31.66 \\ \midrule
RoI-to-CLS / RC               & 63.05 & 35.21 & 62.31 & 30.28 \\
RoI-to-Pooled / Grid          & \textbf{65.19} & 35.88 & \textbf{62.37} & 31.06 \\
RoI-to-Pooled / RC            & 64.91 & \textbf{37.22} & 62.31 & \textbf{32.25} \\
\bottomrule
\end{tabular}
\label{tab:distill feat}
\end{table}


\paragraph{How to distill?}
An effective distillation strategy enables the student model to better absorb the teacher’s knowledge.
While existing works adopt the frozen CLIP vision encoder as the teacher \cite{wu2024clipself, qiu2025scrla, yeo2025atas} to leverage its powerful pretrained representations, and freeze the text encoder during distillation to improve training stability \cite{wu2024clipself}, our findings once again reveal a different perspective for RS-specific adaptation.

We test frozen-teacher distillation on an RS-specific CLIP variant trained with \Lglo~(\cref{tab:distill methods}), where performance drops significantly in most cases compared to the baseline.
Additionally, we evaluate updating the teacher via EMA, a technique widely adopted in self-supervised learning for its training stability \cite{oquab2024dinov}. Nevertheless, the performance remains unsatisfactory.
In contrast, online distillation \cite{jing2024fineclip}, where the teacher and student are trained simultaneously, enables more effective knowledge transfer, as evidenced by its superior accuracy in \cref{tab:distill methods}. 

For the text encoder, \cref{tab: text freeze} shows that freezing it during training limits performance gains. One possible explanation is that the CLIP text encoder lacks sufficient understanding of RS domain language and needs further fine-tuning to better capture domain-specific terms and semantics.

\begin{tcolorbox}[
    colback={rgb,255:red,240;green,249;blue,253},   
    colframe={rgb,255:red,103;green,149;blue,206},  
    rounded corners,        
    boxrule=0.5mm,         
    left=2mm, right=2mm, top=0.5mm, bottom=0.5mm 
]
\textbf{Takeaway 4}: When fine-tuning CLIP to the novel domain of RS with self-distillation, unfreezing both the vision and text encoders results in better performance.
\end{tcolorbox}

\begin{table}[t]
\caption{Comparison of different teacher encoder update strategies for self-distillation.}
\centering
\setlength{\tabcolsep}{2mm}
\begin{tabular}{l|cc|cc}
\toprule
\multirow{2}{*}{Distill method} & \multicolumn{2}{c|}{ViT-B/16} & \multicolumn{2}{c}{ViT-B/32} \\
                        & ZSC   & OVSS   & ZSC   & OVSS \\
\midrule
Baseline                      & 64.13 & 36.34  & 62.22 & 31.66 \\ \midrule
Frozen              & 61.05    & 36.44     & 58.57    & 26.64    \\
EMA                 & 63.16    & 35.84     & 61.60 & 31.52 \\
Online             & \textbf{64.91} & \textbf{37.22}  & \textbf{62.31} & \textbf{32.25} \\
\bottomrule
\end{tabular}
\label{tab:distill methods}
\end{table}

\begin{table}[ht]
\caption{
Comparison of freezing vs. unfreezing the text encoder during self-distillation. Grid cropping is used for local view generation, consistent with \cite{wu2024clipself}.
}
\centering
\setlength{\tabcolsep}{1.5mm}
\begin{tabular}{l|cc|cc}
\toprule
\multirow{2}{*}{Frozen text encoder} & \multicolumn{2}{c|}{ViT-B/16} & \multicolumn{2}{c}{ViT-B/32} \\
                                              & ZSC   & OVSS   & ZSC   & OVSS \\
\midrule
Baseline                                             & 64.13 & \textbf{36.34}  & 62.22 & \textbf{31.66} \\ \midrule
Frozen                       & 62.48 & 34.30  & 59.80 & 29.47    \\
Unfrozen                              & \textbf{65.19} & 35.88  & \textbf{62.37} & 31.06 \\
\bottomrule
\end{tabular}
\label{tab: text freeze}
\end{table}

\paragraph{When to distill?}
We further investigate the effectiveness of patch-to-patch self-distillation under different scenarios, especially with varying data conditions.
In the case of data scale, as shown in \cref{tab: distill scale}, it consistently improves performance in both ZSC and OVSS across nearly all dataset scales and backbones, demonstrating the effectiveness and robustness. 
Notably, the model trained on the 2.5M dataset with self-distillation outperforms the model trained on the 5M dataset without distillation, further demonstrating the benefit of patch-to-patch self-distillation.
In most cases, the highest accuracy is consistently achieved by models trained on the largest dataset combined with self-distillation. 
The only exception is the ViT-B/32 OVSS setting, where performance on the 5M dataset is lower than on the 2.5M dataset.
However, even without distillation (using only \Lglo), the model trained on the 5M dataset still underperforms the one trained on the 2.5M dataset. Therefore, this degradation can be attributed to data-related issues, indicating that when the dataset itself provides poor image-text alignment, applying distillation may harm performance.

We then investigate the impact of self-distillation on the \dn~dataset under two baselines: (1) \Lglo~only, and (2) \Lglo~+ \Lloc. As shown in \cref{tab:ablation-gld}, adding distillation to the \Lglo-only baseline (first vs.\ second row) brings modest improvements overall, though a slight performance drop is observed for OVSS with the ViT-B/32 backbone. This is likely because self-distillation benefits more from ViT-B/16, which offers higher-resolution patch features.
When applied on top of the \Lglo~+~\Lloc~baseline (third vs. fourth row), which is already strong under fine-grained supervision, self-distillation generally has limited effect, bringing only minor improvement with the ViT-B/16 backbone. This suggests that sufficient region-text supervision may reduce the benefits of additional self-distillation.


\begin{table}[ht]
\caption{Comparison of performance with (\cmark) and without (\xmark) self-distillation (\Ldis) on top of image-caption CL (\Lglo) across different data scales and backbones. The hyperparameters are tuned based on the no-distill cases to ensure a fair evaluation of the impact of distillation. Bold and underlined values indicate the best and second-best performance, respectively.}
\centering
\setlength{\tabcolsep}{1.5mm}
\begin{tabular}{l|c|cc|cc} 
\toprule
\multirow{2}{*}{Data scale} & \multirow{2}{*}{Distill} & \multicolumn{2}{c|}{ViT-B/16} & \multicolumn{2}{c}{ViT-B/32} \\
                 &             & ZSC   & OVSS  & ZSC   & OVSS \\
\midrule

\multirow{2}{*}{RS5M-0.5M}   & \xmark     & 61.96 & 36.06 & 60.43 & 31.69 \\  
        & \cmark    & 62.60 & 36.23 & 61.54 & \underline{31.87} \\ \midrule 
\multirow{2}{*}{RS5M-2.5M}   & \xmark     & 64.13 & 36.34 & 62.22 & 31.66 \\
        & \cmark    & \underline{64.91} & \underline{37.22} & \underline{62.31} & \textbf{32.25} \\ \midrule 
\multirow{2}{*}{RS5M-5M}    & \xmark     & 64.76 & 36.57 & 62.16 & 31.53 \\
        & \cmark    & \textbf{65.79} & \textbf{37.30} & \textbf{62.84} & 31.29 \\  
\bottomrule
\end{tabular}
\label{tab: distill scale}
\end{table}

\begin{table}[ht]
\caption{Ablation study of \Ldis, under \Lglo-only and combined \Lglo~and \Lloc~supervision. Training dataset: \dn.}
\centering
\setlength{\tabcolsep}{1.5mm}
\begin{tabular}{ccc|cc|cc}
\toprule
\multirow{2}{*}{\Lglo} & \multirow{2}{*}{\Lloc} & \multirow{2}{*}{\Ldis} & \multicolumn{2}{c|}{ViT-B/16} & \multicolumn{2}{c}{ViT-B/32} \\
       &        &        & ZSC & OVSS & ZSC & OVSS \\
\midrule
\cmark &        &        & 56.63 & 35.19 & 54.28 & \textbf{30.61} \\
\cmark &        & \cmark & 57.39 & \textbf{35.78} & 54.58 & 30.16 \\ \midrule
\cmark & \cmark &        & 58.62 & 35.61 & \textbf{57.88} & 30.28 \\
\cmark & \cmark & \cmark & \textbf{58.79} & 35.63 & 57.83 & 30.20 \\  
\bottomrule
\end{tabular}
\label{tab:ablation-gld}
\end{table}

\begin{table}[ht]
\caption{Comparison of cross-modal retrieval accuracies of different RS-specific CLIP variants across multiple benchmarks. * indicates models trained with in-hold supervision.}
\centering
\fontsize{9pt}{\baselineskip}\selectfont
\setlength{\tabcolsep}{1mm}
\begin{tabular}{l|l|cc|cc|cc}
\toprule
\multirow{2}{*}{Backbone} & \multirow{2}{*}{Method} & \multicolumn{2}{c|}{RSICD} & \multicolumn{2}{c|}{RSITMD} & \multicolumn{2}{c}{UCM} \\
                          &                         & I2T & T2I & I2T & T2I & I2T & T2I \\
\midrule
\multirow{5}{*}{ViT-B/32}
    & CLIP        & 16.3 & 12.8 & 24.7 & 22.4 & 53.2 & 42.5 \\
    & RemoteCLIP*  & \textbf{33.7} & \textbf{28.1} & \textbf{45.7} & \textbf{42.8} & \textbf{77.1} & \textbf{73.5} \\
    & SkyCLIP     & 21.2 & 19.7 & 29.1 & 30.1 & 66.7 & 58.7 \\
    & GeoRSCLIP   & 26.3 & 22.0 & \underline{32.5} & 31.9 & \underline{67.9} & 61.5 \\
    & \mn-s1  & 25.9 & 21.7 & 31.3 & 32.8 & 67.5 & 59.7 \\
    & \mn-s2  & \underline{26.8} & \underline{25.0} & 31.9 & \underline{35.3} & 67.8 & \underline{67.0} \\
\midrule
\multirow{5}{*}{ViT-B/16}
    & CLIP        & 19.8 & 13.4 & 22.9 & 23.4 & 61.4 & 47.5 \\
    & LRSCLIP*     & \textbf{29.7} & \textbf{26.3} & \textbf{39.4} & \underline{33.1} & \textbf{79.5} & \textbf{68.7} \\
    & GeoRSCLIP   & 27.0 & 21.7 & 34.0 & 32.8 & 68.6 & \underline{62.6} \\
    & \mn-s1  & \underline{27.4} & 21.9 & \underline{33.6} & 32.9 & 69.4 & 61.7 \\
    & \mn-s2  & \underline{27.4} & \underline{23.7} & 33.5 & \textbf{34.5} & \underline{70.6} & 62.2 \\

\bottomrule
\end{tabular}
\label{tab:final retrieval}
\end{table}

\begin{table*}[htbp]
\caption{OVSS accuracies across RS benchmarks (mIoU, \%). \textit{G} denotes general-domain models, while \textit{RS} refers to RS-specific models. 
\textit{f.} indicates models specifically designed with fine-grained optimization. 
All models use an input image size of 224, except TIPS (448).
}
\centering
\setlength{\tabcolsep}{1.2mm}
\begin{tabular}{l|l|c|ccccccccc}
\toprule
Backbone & Method & Domain & OEM & LoveDA & Potsdam & Vaihingen & UDD5 & VDD & UAVid & iSAID & Mean \\
\midrule
\multirow{7}{*}{ViT-B/32}
& CLIP \cite{radford2021clip}    & G  & 24.39 & 25.52 & 37.13 & \textbf{25.52} & 35.63 & 35.05 & 28.46 & 6.69 & 27.30 \\
& MetaCLIP \cite{xu2023metaclip} & G  & 24.24 & 26.38 & 35.90 & 20.90 & 31.82 & 30.06 & 29.77 & 7.27 & 25.79 \\
& RemoteCLIP \cite{liu2024remoteclip} & RS & 18.28 & \textbf{33.62} & 21.44 & 21.57 & 29.64 & 29.19 & 17.28 & \underline{14.79} & 23.23 \\
& SkyCLIP \cite{wang2024skyscript}   & RS  & 23.59 & 29.45 & \textbf{40.68} & 23.16 & 34.94 & 30.68 & 28.78 & 3.03 & 26.79 \\
& GeoRSCLIP \cite{zhang2024rs5m} & RS  & \textbf{32.49} & 27.20 & 36.30 & 22.41 & 33.16 & 35.22 & 29.80 & 9.81 & 28.30 \\
& FarSLIP1  & RS / f.  & \underline{28.29} & 30.04 & 36.28 & 22.83 & \textbf{36.80} & \underline{37.23} & \underline{31.42} & \textbf{16.09} & \underline{29.87} \\
& FarSLIP2  & RS / f.  & 26.66 & \underline{30.79} & \underline{40.38} & \underline{24.33} & \underline{36.77} & \textbf{38.54} & \textbf{32.82} & 13.66 & \textbf{30.49} \\
\midrule
ViT-B/14 & TIPS (448) \cite{maninis2024tips} & G / f. & 36.56 & 24.68 & 35.30 & 25.29 & 44.57 & 42.58 & 43.19 & 4.66 & 32.10 \\
\midrule
\multirow{9}{*}{ViT-B/16}
& CLIP \cite{radford2021clip}     & G   & 31.89 & 29.62 & 42.63 & 27.31 & 41.54 & 39.20 & 36.83 & 9.24 & 32.28 \\
& MetaCLIP \cite{xu2023metaclip} & G   & 33.43 & 24.86 & 37.39 & 20.20 & 37.26 & 31.25 & 32.63 & 6.82 & 27.98 \\
& CLIPSelf \cite{wu2024clipself} & G / f.   & 12.61 & 18.36 & 6.27  & 5.51  & 18.44 & 19.99 & 15.80 & 5.22 & 12.78 \\
& FineCLIP \cite{jing2024fineclip}  & G / f.  & 24.01 & 20.82 & 29.62 & 17.41 & 34.53 & 33.86 & 26.69 & 6.54 & 24.19 \\
& COSMOS \cite{kim2025cosmos}  & G / f.  & 21.90 & 26.16 & 31.23 & 21.57 & 35.90 & 30.86 & 26.38 & 5.91 & 24.99 \\
& LRSCLIP \cite{chen2025lrsclip}  & RS  & 30.75 & \textbf{34.61} & 43.98 & \textbf{26.68} & 44.24 & 37.90 & 37.57 & 10.11 & 33.23 \\
& GeoRSCLIP \cite{zhang2024rs5m} & RS & \underline{35.26} & 34.10 & 45.52 & \underline{23.35} & 43.22 & 40.57 & 39.13 & 13.56 & 34.34 \\
& FarSLIP1 & RS / f.  & \textbf{35.89} & 33.59 & \underline{46.50} & 22.34 & \underline{44.60} & \underline{40.63} & \textbf{40.23} & \textbf{19.76} & \textbf{35.44} \\
& FarSLIP2 & RS / f.  & 34.39 & \underline{34.23} & \textbf{47.48} & 22.81 & \textbf{44.83} & \textbf{40.82} & \underline{40.03} & \underline{18.67} & \underline{35.41} \\
\bottomrule
\end{tabular}
\label{tab:final ovss}
\end{table*}

\begin{table*}[ht]
\caption{Comparison of ZSC accuracies (Top-1 acc., \%) of different RS-specific CLIP variants across multiple benchmarks.}
\centering
\setlength{\tabcolsep}{1.2mm}
\begin{tabular}{l|l|ccccccccc}
\toprule
{Backbone} & {Method} & {SkyScript} & {AID} & {EuroSAT} & {FMoW} & {MillionAID} & {PatternNet} & {NWPU} & {RSICB} & {Mean} \\
\midrule
\multirow{6}{*}{ViT-B/32} 
    & CLIP  \cite{radford2021clip}           & 32.37 & 58.00 & 22.00 & 15.61 & 51.62 & 52.97 & 54.82 & 29.92 & 39.66 \\
    & RemoteCLIP \cite{liu2024remoteclip}     & 27.67 & \textbf{88.10} & 28.52 & 11.15 & 45.50 & 55.95 & 69.10 & 43.91 & 46.24 \\
    & SkyCLIP \cite{wang2024skyscript}        & \underline{52.92} & 70.90 & 33.33 & 19.23 & 62.72 & 72.19 & 66.63 & 46.20 & 53.02 \\
    & GeoRSCLIP \cite{zhang2024rs5m}      & 51.92 & 69.75 & 41.33 & \textbf{37.14} & \underline{71.60} & \underline{77.00} & \underline{69.90} & \underline{46.93} & 58.20 \\
    & \mn-s1      & 51.51 & \underline{73.80} & \underline{49.26} & \underline{34.49} & 69.42 & 74.69 & \textbf{69.93} & 46.00 & \underline{58.64} \\
    & \mn-s2      & \textbf{53.79} & 73.30 & \textbf{49.30} & 30.12 & \textbf{72.05} & \textbf{80.45} & 67.77 & \textbf{54.17} & \textbf{60.12} \\

\midrule
\multirow{5}{*}{ViT-B/16} 
    & CLIP \cite{radford2021clip}           & 42.60 & 65.85 & 32.78 & 19.85 & 53.37 & 63.82 & 62.25 & 33.56 & 46.76 \\
    & LRSCLIP \cite{chen2025lrsclip}        & 55.73 & 69.75 & 44.52 & 22.66 & \underline{67.76} & 77.89 & \textbf{74.28} & \textbf{54.75} & 58.42 \\
    & GeoRSCLIP  \cite{zhang2024rs5m}     & 55.33 & 72.15 & 54.59 & 34.50 & 67.27 & 78.75 & 72.32 & 52.75 & 60.96 \\
    & \mn-s1      & \underline{56.19} & \underline{73.30} & \textbf{59.81} & \underline{34.60} & 66.82 & \underline{78.85} & \underline{72.72} & 52.80 & \underline{61.89} \\
    & \mn-s2      & \textbf{57.98} & \textbf{73.95} & \underline{55.52} & \textbf{35.86} & \textbf{68.04} & \textbf{80.82} & 71.49 & \underline{54.24} & \textbf{62.24} \\
\bottomrule
\end{tabular}
\label{tab:final zsc}
\end{table*}

\section{\mn: A Family of RS VLFM}
\label{sec:results}

Based on the above insights, we propose a new RS-specific VLFM named \mn~(\cref{fig:architecture}), trained in a two-stage manner. 
In stage one (\mn-s1), the model is trained with both \Lglo~and \Ldis, using the RS5M dataset for ViT-B/16 and RS5M-2.5M subset for ViT-B/32. We adopt patch-to-patch self-distillation with random crops to generate local views, and unfreeze both the teacher and student vision encoders as well as the text encoder.
In stage two (\mn-s2), the model is further fine-tuned on the \dn~dataset using \Lglo~and \Lloc. The entire training process employs full-parameter fine-tuning, initialized from OpenAI CLIP.
Detailed  settings are provided in Appendix.

We compare \mn~with several RS-specific VLFMs (when their models are publicly available) and general-domain VLFMs equipped with fine-grained vision-language alignment capabilities.
\cref{tab:final ovss} and \cref{tab:final zsc} report OVSS and ZSC results, respectively. \mn-s1 achieves competitive or best performance across all benchmarks and outperforms other VLFMs in mean accuracy, demonstrating its strength in both global and fine-grained understanding. \mn-s2, further fine-tuned on our proposed \dn~dataset (which is fully out-of-distribution relative to the adopted benchmarks), brings additional improvements and achieves the highest overall performance.
We further evaluate cross-modal retrieval performance of \mn, as shown in \cref{tab:final retrieval}. Excluding models trained on hold-in data, \mn~achieves either SOTA or competitive results.

\section{Conclusion}
\label{sec:conclusion}

This study explores fine-grained adaption of CLIP for RS domain. With our proposed \dn~dataset, constructed from public resources, we demonstrate that current RS image-text data generation paradigms often underutilize valuable region-level information, which can in fact provide effective supervision for both global- and local-level vision-language alignment. Our findings offer practical guidance for future RS vision-language data construction.

Furthermore, we conduct an in-depth investigation of existing fine-grained CLIP fine-tuning strategies and show that current region-language alignment approaches can undermine semantic coherence, thereby degrading performance on fine-grained RS interpretation tasks.
By applying appropriate training objectives, including patch-to-patch self-distillation and basic CLS token-based region-category CL, we achieve a SOTA VLFM specialized for RS. We hope that this study can provide valuable insights for fine-grained adaptation of CLIP to downstream domains.



{
    \small
    \bibliographystyle{ieeenat_fullname}
    \bibliography{main}

\begin{thebibliography}{76}
\providecommand{\natexlab}[1]{#1}
\providecommand{\url}[1]{\texttt{#1}}
\expandafter\ifx\csname urlstyle\endcsname\relax
  \providecommand{\doi}[1]{doi: #1}\else
  \providecommand{\doi}{doi: \begingroup \urlstyle{rm}\Url}\fi

\bibitem[Asokan et~al.(2025)Asokan, Wu, and Albreiki]{asokan2025finelip}
Mothilal Asokan, Kebin Wu, and Fatima Albreiki.
\newblock Finelip: Extending clip's reach via fine-grained alignment with
  longer text inputs.
\newblock In \emph{Proceedings of the Computer Vision and Pattern Recognition
  Conference}, pages 14495--14504, 2025.

\bibitem[Barsellotti et~al.(2025)Barsellotti, Bianchi, Messina, Carrara,
  Cornia, Baraldi, Falchi, and Cucchiara]{barsellotti2025talkingtoDINO}
Luca Barsellotti, Lorenzo Bianchi, Nicola Messina, Fabio Carrara, Marcella
  Cornia, Lorenzo Baraldi, Fabrizio Falchi, and Rita Cucchiara.
\newblock Talking to dino: Bridging self-supervised vision backbones with
  language for open-vocabulary segmentation.
\newblock In \emph{Proceedings of the IEEE/CVF International Conference on
  Computer Vision}, pages 22025--22035, 2025.

\bibitem[Bica et~al.(2024)Bica, Ili\'{c}, Bauer, Erdogan, Bo\v{s}njak,
  Kaplanis, Gritsenko, Minderer, Blundell, Pa\c{s}canu, and
  Mitrovi\'{c}]{sparc}
Ioana Bica, Anastasija Ili\'{c}, Matthias Bauer, Goker Erdogan, Matko
  Bo\v{s}njak, Christos Kaplanis, Alexey~A. Gritsenko, Matthias Minderer,
  Charles Blundell, Razvan Pa\c{s}canu, and Jovana Mitrovi\'{c}.
\newblock Improving fine-grained understanding in image-text pre-training.
\newblock In \emph{Proceedings of the 41st International Conference on Machine
  Learning}, 2024.

\bibitem[Cai et~al.(2025)Cai, Jin, Hou, Guo, Wu, and Yang]{cai2025vdd}
Wenxiao Cai, Ke Jin, Jinyan Hou, Cong Guo, Letian Wu, and Wankou Yang.
\newblock Vdd: Varied drone dataset for semantic segmentation.
\newblock \emph{Journal of Visual Communication and Image Representation},
  109:\penalty0 104429, 2025.

\bibitem[Chen et~al.(2024)Chen, Lai, Zhang, Wang, Eichner, You, Cao, Zhang,
  Yang, and Gan]{chen2024cloc}
Hong-You Chen, Zhengfeng Lai, Haotian Zhang, Xinze Wang, Marcin Eichner, Keen
  You, Meng Cao, Bowen Zhang, Yinfei Yang, and Zhe Gan.
\newblock Contrastive localized language-image pre-training.
\newblock \emph{arXiv preprint arXiv:2410.02746}, 2024.

\bibitem[Chen et~al.(2025)Chen, Chen, Deng, Chen, Feng, Xi, Liu, Li, and
  Meng]{chen2025lrsclip}
Weizhi Chen, Jingbo Chen, Yupeng Deng, Jiansheng Chen, Yuman Feng, Zhihao Xi,
  Diyou Liu, Kai Li, and Yu Meng.
\newblock Lrsclip: A vision-language foundation model for aligning remote
  sensing image with longer text.
\newblock \emph{arXiv preprint arXiv:2503.19311}, 2025.

\bibitem[Chen et~al.(2018)Chen, Wang, Lu, Chen, and Wang]{chen2018udd}
Yu Chen, Yao Wang, Peng Lu, Yisong Chen, and Guoping Wang.
\newblock Large-scale structure from motion with semantic constraints of aerial
  images.
\newblock In \emph{Chinese Conference on Pattern Recognition and Computer
  Vision (PRCV)}, pages 347--359. Springer, 2018.

\bibitem[Cheng et~al.(2014)Cheng, Han, Zhou, and Guo]{cheng2014NWPU-VHR}
Gong Cheng, Junwei Han, Peicheng Zhou, and Lei Guo.
\newblock Multi-class geospatial object detection and geographic image
  classification based on collection of part detectors.
\newblock \emph{ISPRS Journal of Photogrammetry and Remote Sensing},
  98:\penalty0 119--132, 2014.

\bibitem[Cheng et~al.(2017)Cheng, Han, and Lu]{cheng2017RESISC45}
Gong Cheng, Junwei Han, and Xiaoqiang Lu.
\newblock Remote sensing image scene classification: Benchmark and state of the
  art.
\newblock \emph{Proceedings of the IEEE}, 105\penalty0 (10):\penalty0
  1865--1883, 2017.

\bibitem[Choi et~al.(2025)Choi, Jang, and Eom]{choi2025goal}
Hyungyu Choi, Young~Kyun Jang, and Chanho Eom.
\newblock Goal: Global-local object alignment learning.
\newblock In \emph{Proceedings of the Computer Vision and Pattern Recognition
  Conference}, pages 4070--4079, 2025.

\bibitem[Christie et~al.(2018)Christie, Fendley, Wilson, and
  Mukherjee]{christie2018fmow}
Gordon Christie, Neil Fendley, James Wilson, and Ryan Mukherjee.
\newblock Functional map of the world.
\newblock In \emph{Proceedings of the IEEE Conference on Computer Vision and
  Pattern Recognition}, pages 6172--6180, 2018.

\bibitem[Davies and Bouldin(2009)]{davies2009DBI}
David~L Davies and Donald~W Bouldin.
\newblock A cluster separation measure.
\newblock \emph{IEEE transactions on pattern analysis and machine
  intelligence}, \penalty0 (2):\penalty0 224--227, 2009.

\bibitem[Fu et~al.(2024)Fu, Hamilton, Brandt, Feldmann, Zhang, and
  Freeman]{fu2024featup}
Stephanie Fu, Mark Hamilton, Laura~E. Brandt, Axel Feldmann, Zhoutong Zhang,
  and William~T. Freeman.
\newblock Featup: A model-agnostic framework for features at any resolution.
\newblock In \emph{The Twelfth International Conference on Learning
  Representations}, 2024.

\bibitem[Goyal et~al.(2023)Goyal, Kumar, Garg, Kolter, and
  Raghunathan]{goyal2023finetune}
Sachin Goyal, Ananya Kumar, Sankalp Garg, Zico Kolter, and Aditi Raghunathan.
\newblock Finetune like you pretrain: Improved finetuning of zero-shot vision
  models.
\newblock In \emph{Proceedings of the IEEE/CVF Conference on Computer Vision
  and Pattern Recognition}, pages 19338--19347, 2023.

\bibitem[He et~al.(2017)He, Gkioxari, Doll{\'a}r, and Girshick]{he2017maskrcnn}
Kaiming He, Georgia Gkioxari, Piotr Doll{\'a}r, and Ross Girshick.
\newblock Mask r-cnn.
\newblock In \emph{Proceedings of the IEEE international conference on computer
  vision}, pages 2961--2969, 2017.

\bibitem[Helber et~al.(2019)Helber, Bischke, Dengel, and
  Borth]{helber2019eurosat}
Patrick Helber, Benjamin Bischke, Andreas Dengel, and Damian Borth.
\newblock Eurosat: A novel dataset and deep learning benchmark for land use and
  land cover classification.
\newblock \emph{IEEE Journal of Selected Topics in Applied Earth Observations
  and Remote Sensing}, 12\penalty0 (7):\penalty0 2217--2226, 2019.

\bibitem[Hsieh et~al.(2017)Hsieh, Lin, and Hsu]{hsieh2017carpk}
Meng-Ru Hsieh, Yen-Liang Lin, and Winston~H Hsu.
\newblock Drone-based object counting by spatially regularized regional
  proposal network.
\newblock In \emph{Proceedings of the IEEE international conference on computer
  vision}, pages 4145--4153, 2017.

\bibitem[{ISPRS}(2016{\natexlab{a}})]{Potsdam}
{ISPRS}.
\newblock {2D Semantic Labeling Contest - Potsdam Dataset}.
\newblock
  \url{https://www.isprs.org/resources/datasets/benchmarks/UrbanSemLab/2d-sem-label-potsdam.aspx},
  2016{\natexlab{a}}.
\newblock Accessed: 2025-07-31.

\bibitem[{ISPRS}(2016{\natexlab{b}})]{Vaihingen}
{ISPRS}.
\newblock {2D Semantic Labeling Contest - Vaihingen Dataset}.
\newblock
  \url{https://www.isprs.org/resources/datasets/benchmarks/UrbanSemLab/2d-sem-label-vaihingen.aspx},
  2016{\natexlab{b}}.
\newblock Accessed: 2025-07-31.

\bibitem[Jing et~al.(2024)Jing, He, Luo, Fei, Wei, Zhao, Lu,
  et~al.]{jing2024fineclip}
Dong Jing, Xiaolong He, Yutian Luo, Nanyi Fei, Wei Wei, Huiwen Zhao, Zhiwu Lu,
  et~al.
\newblock Fineclip: Self-distilled region-based clip for better fine-grained
  understanding.
\newblock \emph{Advances in Neural Information Processing Systems},
  37:\penalty0 27896--27918, 2024.

\bibitem[Khosla et~al.(2020)Khosla, Teterwak, Wang, Sarna, Tian, Isola,
  Maschinot, Liu, and Krishnan]{khosla2020SupCon}
Prannay Khosla, Piotr Teterwak, Chen Wang, Aaron Sarna, Yonglong Tian, Phillip
  Isola, Aaron Maschinot, Ce Liu, and Dilip Krishnan.
\newblock Supervised contrastive learning.
\newblock \emph{Advances in neural information processing systems},
  33:\penalty0 18661--18673, 2020.

\bibitem[Kim et~al.(2025)Kim, Xiao, Georgescu, Alaniz, and
  Akata]{kim2025cosmos}
Sanghwan Kim, Rui Xiao, Mariana-Iuliana Georgescu, Stephan Alaniz, and Zeynep
  Akata.
\newblock Cosmos: Cross-modality self-distillation for vision language
  pre-training.
\newblock In \emph{Proceedings of the Computer Vision and Pattern Recognition
  Conference}, pages 14690--14700, 2025.

\bibitem[Kumar et~al.(2022)Kumar, Raghunathan, Jones, Ma, and
  Liang]{kumar2022fine}
Ananya Kumar, Aditi Raghunathan, Robbie Jones, Tengyu Ma, and Percy Liang.
\newblock Fine-tuning can distort pretrained features and underperform
  out-of-distribution.
\newblock \emph{arXiv preprint arXiv:2202.10054}, 2022.

\bibitem[Lam et~al.(2018)Lam, Kuzma, McGee, Dooley, Laielli, Klaric, Bulatov,
  and McCord]{lam2018xview}
Darius Lam, Richard Kuzma, Kevin McGee, Samuel Dooley, Michael Laielli, Matthew
  Klaric, Yaroslav Bulatov, and Brendan McCord.
\newblock xview: Objects in context in overhead imagery.
\newblock \emph{arXiv preprint arXiv:1802.07856}, 2018.

\bibitem[Lan et~al.(2024{\natexlab{a}})Lan, Chen, Ke, Wang, Feng, and
  Zhang]{lan2024clearclip}
Mengcheng Lan, Chaofeng Chen, Yiping Ke, Xinjiang Wang, Litong Feng, and Wayne
  Zhang.
\newblock Clearclip: Decomposing clip representations for dense vision-language
  inference.
\newblock In \emph{European Conference on Computer Vision}, pages 143--160.
  Springer, 2024{\natexlab{a}}.

\bibitem[Lan et~al.(2024{\natexlab{b}})Lan, Chen, Ke, Wang, Feng, and
  Zhang]{lan2024proxyclip}
Mengcheng Lan, Chaofeng Chen, Yiping Ke, Xinjiang Wang, Litong Feng, and Wayne
  Zhang.
\newblock Proxyclip: Proxy attention improves clip for open-vocabulary
  segmentation.
\newblock In \emph{European Conference on Computer Vision}, pages 70--88.
  Springer, 2024{\natexlab{b}}.

\bibitem[Li et~al.(2020{\natexlab{a}})Li, Dou, Tao, Wu, Chen, Peng, Deng, and
  Zhao]{li2020rsicb}
Haifeng Li, Xin Dou, Chao Tao, Zhixiang Wu, Jie Chen, Jian Peng, Min Deng, and
  Ling Zhao.
\newblock Rsi-cb: A large-scale remote sensing image classification benchmark
  using crowdsourced data.
\newblock \emph{Sensors}, 20\penalty0 (6):\penalty0 1594, 2020{\natexlab{a}}.

\bibitem[Li et~al.(2020{\natexlab{b}})Li, Wan, Cheng, Meng, and
  Han]{li2020DIOR}
Ke Li, Gang Wan, Gong Cheng, Liqiu Meng, and Junwei Han.
\newblock Object detection in optical remote sensing images: A survey and a new
  benchmark.
\newblock \emph{ISPRS journal of photogrammetry and remote sensing},
  159:\penalty0 296--307, 2020{\natexlab{b}}.

\bibitem[Li et~al.(2025{\natexlab{a}})Li, Liu, Cao, Bai, Zhou, Meng, and
  Wang]{li2025segearth}
Kaiyu Li, Ruixun Liu, Xiangyong Cao, Xueru Bai, Feng Zhou, Deyu Meng, and Zhi
  Wang.
\newblock Segearth-ov: Towards training-free open-vocabulary segmentation for
  remote sensing images.
\newblock In \emph{Proceedings of the Computer Vision and Pattern Recognition
  Conference}, pages 10545--10556, 2025{\natexlab{a}}.

\bibitem[Li et~al.(2025{\natexlab{b}})Li, Muhtar, Gu, He, Zhang, Xiao, He, and
  Zhu]{li2025lhrs}
Zhenshi Li, Dilxat Muhtar, Feng Gu, Yanglangxing He, Xueliang Zhang, Pengfeng
  Xiao, Guangjun He, and Xiaoxiang Zhu.
\newblock Lhrs-bot-nova: Improved multimodal large language model for remote
  sensing vision-language interpretation.
\newblock \emph{ISPRS Journal of Photogrammetry and Remote Sensing},
  227:\penalty0 539--550, 2025{\natexlab{b}}.

\bibitem[Liu et~al.(2024)Liu, Chen, Guan, Zhou, Zhu, Ye, Fu, and
  Zhou]{liu2024remoteclip}
Fan Liu, Delong Chen, Zhangqingyun Guan, Xiaocong Zhou, Jiale Zhu, Qiaolin Ye,
  Liyong Fu, and Jun Zhou.
\newblock Remoteclip: A vision language foundation model for remote sensing.
\newblock \emph{IEEE Transactions on Geoscience and Remote Sensing},
  62:\penalty0 1--16, 2024.

\bibitem[Liu et~al.(2017)Liu, Yuan, Weng, and Yang]{liu2017HRSC}
Zikun Liu, Liu Yuan, Lubin Weng, and Yiping Yang.
\newblock A high resolution optical satellite image dataset for ship
  recognition and some new baselines.
\newblock In \emph{International conference on pattern recognition applications
  and methods}, pages 324--331. SciTePress, 2017.

\bibitem[Long et~al.(2017)Long, Gong, Xiao, and Liu]{long2017rsod}
Yang Long, Yiping Gong, Zhifeng Xiao, and Qing Liu.
\newblock Accurate object localization in remote sensing images based on
  convolutional neural networks.
\newblock \emph{IEEE Transactions on Geoscience and Remote Sensing},
  55\penalty0 (5):\penalty0 2486--2498, 2017.

\bibitem[Long et~al.(2021)Long, Xia, Li, Yang, Yang, Zhu, Zhang, and
  Li]{long2021millionaid}
Yang Long, Gui-Song Xia, Shengyang Li, Wen Yang, Michael~Ying Yang, Xiao~Xiang
  Zhu, Liangpei Zhang, and Deren Li.
\newblock On creating benchmark dataset for aerial image interpretation:
  Reviews, guidances, and million-aid.
\newblock \emph{IEEE Journal of selected topics in applied earth observations
  and remote sensing}, 14:\penalty0 4205--4230, 2021.

\bibitem[Lu et~al.(2017)Lu, Wang, Zheng, and Li]{lu2017rsicd}
Xiaoqiang Lu, Binqiang Wang, Xiangtao Zheng, and Xuelong Li.
\newblock Exploring models and data for remote sensing image caption
  generation.
\newblock \emph{IEEE Transactions on Geoscience and Remote Sensing},
  56\penalty0 (4):\penalty0 2183--2195, 2017.

\bibitem[Lyu et~al.(2020)Lyu, Vosselman, Xia, Yilmaz, and Yang]{lyu2020uavid}
Ye Lyu, George Vosselman, Gui-Song Xia, Alper Yilmaz, and Michael~Ying Yang.
\newblock Uavid: A semantic segmentation dataset for uav imagery.
\newblock \emph{ISPRS journal of photogrammetry and remote sensing},
  165:\penalty0 108--119, 2020.

\bibitem[Maninis et~al.(2024)Maninis, Chen, Ghosh, Karpur, Chen, Xia, Cao,
  Salz, Han, Dlabal, et~al.]{maninis2024tips}
Kevis-Kokitsi Maninis, Kaifeng Chen, Soham Ghosh, Arjun Karpur, Koert Chen, Ye
  Xia, Bingyi Cao, Daniel Salz, Guangxing Han, Jan Dlabal, et~al.
\newblock Tips: Text-image pretraining with spatial awareness.
\newblock \emph{arXiv preprint arXiv:2410.16512}, 2024.

\bibitem[Muhtar et~al.(2024)Muhtar, Li, Gu, Zhang, and Xiao]{muhtar2024lhrs}
Dilxat Muhtar, Zhenshi Li, Feng Gu, Xueliang Zhang, and Pengfeng Xiao.
\newblock Lhrs-bot: Empowering remote sensing with vgi-enhanced large
  multimodal language model.
\newblock In \emph{European Conference on Computer Vision}, pages 440--457.
  Springer, 2024.

\bibitem[Mukhoti et~al.(2023{\natexlab{a}})Mukhoti, Gal, Torr, and
  Dokania]{mukhoti2023fine}
Jishnu Mukhoti, Yarin Gal, Philip~HS Torr, and Puneet~K Dokania.
\newblock Fine-tuning can cripple your foundation model; preserving features
  may be the solution.
\newblock \emph{arXiv preprint arXiv:2308.13320}, 2023{\natexlab{a}}.

\bibitem[Mukhoti et~al.(2023{\natexlab{b}})Mukhoti, Lin, Poursaeed, Wang, Shah,
  Torr, and Lim]{mukhoti2023pacl}
Jishnu Mukhoti, Tsung-Yu Lin, Omid Poursaeed, Rui Wang, Ashish Shah, Philip~HS
  Torr, and Ser-Nam Lim.
\newblock Open vocabulary semantic segmentation with patch aligned contrastive
  learning.
\newblock In \emph{Proceedings of the IEEE/CVF Conference on Computer Vision
  and Pattern Recognition}, pages 19413--19423, 2023{\natexlab{b}}.

\bibitem[Naeem et~al.(2024)Naeem, Xian, Zhai, Hoyer, Van~Gool, and
  Tombari]{naeem2024silc}
Muhammad~Ferjad Naeem, Yongqin Xian, Xiaohua Zhai, Lukas Hoyer, Luc Van~Gool,
  and Federico Tombari.
\newblock Silc: Improving vision language pretraining with self-distillation.
\newblock In \emph{European Conference on Computer Vision}, pages 38--55.
  Springer, 2024.

\bibitem[Oquab et~al.(2024)Oquab, Darcet, Moutakanni, Vo, Szafraniec, Khalidov,
  Fernandez, HAZIZA, Massa, El-Nouby, Assran, Ballas, Galuba, Howes, Huang, Li,
  Misra, Rabbat, Sharma, Synnaeve, Xu, Jegou, Mairal, Labatut, Joulin, and
  Bojanowski]{oquab2024dinov}
Maxime Oquab, Timoth{\'e}e Darcet, Th{\'e}o Moutakanni, Huy~V. Vo, Marc
  Szafraniec, Vasil Khalidov, Pierre Fernandez, Daniel HAZIZA, Francisco Massa,
  Alaaeldin El-Nouby, Mido Assran, Nicolas Ballas, Wojciech Galuba, Russell
  Howes, Po-Yao Huang, Shang-Wen Li, Ishan Misra, Michael Rabbat, Vasu Sharma,
  Gabriel Synnaeve, Hu Xu, Herve Jegou, Julien Mairal, Patrick Labatut, Armand
  Joulin, and Piotr Bojanowski.
\newblock {DINO}v2: Learning robust visual features without supervision.
\newblock \emph{Transactions on Machine Learning Research}, 2024.

\bibitem[Pan et~al.(2025)Pan, Liu, Fu, Ma, Li, Paudel, Van~Gool, and
  Huang]{pan2025LAE}
Jiancheng Pan, Yanxing Liu, Yuqian Fu, Muyuan Ma, Jiahao Li, Danda~Pani Paudel,
  Luc Van~Gool, and Xiaomeng Huang.
\newblock Locate anything on earth: Advancing open-vocabulary object detection
  for remote sensing community.
\newblock In \emph{Proceedings of the AAAI Conference on Artificial
  Intelligence}, pages 6281--6289, 2025.

\bibitem[Qiu et~al.(2025)Qiu, Wu, Ke, Bai, and Zhang]{qiu2025scrla}
Congpei Qiu, Yanhao Wu, Wei Ke, Xiuxiu Bai, and Tong Zhang.
\newblock Refining {CLIP}'s spatial awareness: A visual-centric perspective.
\newblock In \emph{The Thirteenth International Conference on Learning
  Representations}, 2025.

\bibitem[Radford et~al.(2021)Radford, Kim, Hallacy, Ramesh, Goh, Agarwal,
  Sastry, Askell, Mishkin, Clark, et~al.]{radford2021clip}
Alec Radford, Jong~Wook Kim, Chris Hallacy, Aditya Ramesh, Gabriel Goh,
  Sandhini Agarwal, Girish Sastry, Amanda Askell, Pamela Mishkin, Jack Clark,
  et~al.
\newblock Learning transferable visual models from natural language
  supervision.
\newblock In \emph{International conference on machine learning}, pages
  8748--8763. PmLR, 2021.

\bibitem[Sun et~al.(2022)Sun, Wang, Yan, Xu, Wang, Diao, Chen, Li, Feng, Xu,
  et~al.]{sun2022fair1m}
Xian Sun, Peijin Wang, Zhiyuan Yan, Feng Xu, Ruiping Wang, Wenhui Diao, Jin
  Chen, Jihao Li, Yingchao Feng, Tao Xu, et~al.
\newblock Fair1m: A benchmark dataset for fine-grained object recognition in
  high-resolution remote sensing imagery.
\newblock \emph{ISPRS Journal of Photogrammetry and Remote Sensing},
  184:\penalty0 116--130, 2022.

\bibitem[Tian et~al.(2023)Tian, Fan, Isola, Chang, and
  Krishnan]{tian2023stablerep}
Yonglong Tian, Lijie Fan, Phillip Isola, Huiwen Chang, and Dilip Krishnan.
\newblock Stablerep: Synthetic images from text-to-image models make strong
  visual representation learners.
\newblock \emph{Advances in Neural Information Processing Systems},
  36:\penalty0 48382--48402, 2023.

\bibitem[Tschannen et~al.(2025)Tschannen, Gritsenko, Wang, Naeem,
  Alabdulmohsin, Parthasarathy, Evans, Beyer, Xia, Mustafa,
  et~al.]{tschannen2025siglip2}
Michael Tschannen, Alexey Gritsenko, Xiao Wang, Muhammad~Ferjad Naeem, Ibrahim
  Alabdulmohsin, Nikhil Parthasarathy, Talfan Evans, Lucas Beyer, Ye Xia, Basil
  Mustafa, et~al.
\newblock Siglip 2: Multilingual vision-language encoders with improved
  semantic understanding, localization, and dense features.
\newblock \emph{arXiv preprint arXiv:2502.14786}, 2025.

\bibitem[Vujasinovi{\'c} et~al.(2020)Vujasinovi{\'c}, Becker, Breuer,
  Bullinger, Scherer-Negenborn, and Arens]{vujasinovic2020auair}
St{\'e}phane Vujasinovi{\'c}, Stefan Becker, Timo Breuer, Sebastian Bullinger,
  Norbert Scherer-Negenborn, and Michael Arens.
\newblock Integration of the 3d environment for uav onboard visual object
  tracking.
\newblock \emph{Applied Sciences}, 10\penalty0 (21):\penalty0 7622, 2020.

\bibitem[Wang et~al.(2024{\natexlab{a}})Wang, Mei, and Yuille]{wang2024sclip}
Feng Wang, Jieru Mei, and Alan Yuille.
\newblock Sclip: Rethinking self-attention for dense vision-language inference.
\newblock In \emph{European Conference on Computer Vision}, pages 315--332.
  Springer, 2024{\natexlab{a}}.

\bibitem[Wang et~al.(2021)Wang, Zheng, Ma, Lu, and Zhong]{wang2021loveda}
Junjue Wang, Zhuo Zheng, Ailong Ma, Xiaoyan Lu, and Yanfei Zhong.
\newblock Loveda: A remote sensing land-cover dataset for domain adaptive
  semantic segmentation.
\newblock \emph{arXiv preprint arXiv:2110.08733}, 2021.

\bibitem[Wang et~al.(2024{\natexlab{b}})Wang, Prabha, Huang, Wu, and
  Rajagopal]{wang2024skyscript}
Zhecheng Wang, Rajanie Prabha, Tianyuan Huang, Jiajun Wu, and Ram Rajagopal.
\newblock Skyscript: A large and semantically diverse vision-language dataset
  for remote sensing.
\newblock In \emph{Proceedings of the AAAI Conference on Artificial
  Intelligence}, pages 5805--5813, 2024{\natexlab{b}}.

\bibitem[Waqas~Zamir et~al.(2019)Waqas~Zamir, Arora, Gupta, Khan, Sun,
  Shahbaz~Khan, Zhu, Shao, Xia, and Bai]{waqas2019isaid}
Syed Waqas~Zamir, Aditya Arora, Akshita Gupta, Salman Khan, Guolei Sun, Fahad
  Shahbaz~Khan, Fan Zhu, Ling Shao, Gui-Song Xia, and Xiang Bai.
\newblock isaid: A large-scale dataset for instance segmentation in aerial
  images.
\newblock In \emph{Proceedings of the IEEE/CVF conference on computer vision
  and pattern recognition workshops}, pages 28--37, 2019.

\bibitem[Wu et~al.(2024)Wu, Zhang, Xu, Jin, Li, Liu, and Loy]{wu2024clipself}
Size Wu, Wenwei Zhang, Lumin Xu, Sheng Jin, Xiangtai Li, Wentao Liu, and
  Chen~Change Loy.
\newblock {CLIPS}elf: Vision transformer distills itself for open-vocabulary
  dense prediction.
\newblock In \emph{The Twelfth International Conference on Learning
  Representations}, 2024.

\bibitem[Wysocza{\'n}ska et~al.(2024)Wysocza{\'n}ska, Sim{\'e}oni,
  Ramamonjisoa, Bursuc, Trzci{\'n}ski, and
  P{\'e}rez]{wysoczanska2024clipDINOise}
Monika Wysocza{\'n}ska, Oriane Sim{\'e}oni, Micha{\"e}l Ramamonjisoa, Andrei
  Bursuc, Tomasz Trzci{\'n}ski, and Patrick P{\'e}rez.
\newblock Clip-dinoiser: Teaching clip a few dino tricks for open-vocabulary
  semantic segmentation.
\newblock In \emph{European Conference on Computer Vision}, pages 320--337.
  Springer, 2024.

\bibitem[Xia et~al.(2017)Xia, Hu, Hu, Shi, Bai, Zhong, Zhang, and
  Lu]{xia2017aid}
Gui-Song Xia, Jingwen Hu, Fan Hu, Baoguang Shi, Xiang Bai, Yanfei Zhong,
  Liangpei Zhang, and Xiaoqiang Lu.
\newblock Aid: A benchmark data set for performance evaluation of aerial scene
  classification.
\newblock \emph{IEEE Transactions on Geoscience and Remote Sensing},
  55\penalty0 (7):\penalty0 3965--3981, 2017.

\bibitem[Xia et~al.(2018)Xia, Bai, Ding, Zhu, Belongie, Luo, Datcu, Pelillo,
  and Zhang]{xia2018dota}
Gui-Song Xia, Xiang Bai, Jian Ding, Zhen Zhu, Serge Belongie, Jiebo Luo, Mihai
  Datcu, Marcello Pelillo, and Liangpei Zhang.
\newblock Dota: A large-scale dataset for object detection in aerial images.
\newblock In \emph{Proceedings of the IEEE conference on computer vision and
  pattern recognition}, pages 3974--3983, 2018.

\bibitem[Xia et~al.(2023)Xia, Yokoya, Adriano, and
  Broni-Bediako]{xia2023openearthmap}
Junshi Xia, Naoto Yokoya, Bruno Adriano, and Clifford Broni-Bediako.
\newblock Openearthmap: A benchmark dataset for global high-resolution land
  cover mapping.
\newblock In \emph{Proceedings of the IEEE/CVF Winter Conference on
  Applications of Computer Vision}, pages 6254--6264, 2023.

\bibitem[Xiao et~al.(2025)Xiao, Kim, Georgescu, Akata, and
  Alaniz]{xiao2025flair}
Rui Xiao, Sanghwan Kim, Mariana-Iuliana Georgescu, Zeynep Akata, and Stephan
  Alaniz.
\newblock Flair: Vlm with fine-grained language-informed image representations.
\newblock In \emph{Proceedings of the Computer Vision and Pattern Recognition
  Conference}, pages 24884--24894, 2025.

\bibitem[Xie et~al.(2025)Xie, Wang, Kong, Li, Liang, Zhang, Leng, and
  Yin]{xie2025FG-CLIP}
Chunyu Xie, Bin Wang, Fanjing Kong, Jincheng Li, Dawei Liang, Gengshen Zhang,
  Dawei Leng, and Yuhui Yin.
\newblock Fg-clip: Fine-grained visual and textual alignment.
\newblock \emph{arXiv preprint arXiv:2505.05071}, 2025.

\bibitem[Xiong et~al.(2025)Xiong, Wang, Yu, Stewart, Zhao, Lehmann, Dujardin,
  Yuan, Ghamisi, and Zhu]{xiong2025geolangbind}
Zhitong Xiong, Yi Wang, Weikang Yu, Adam~J Stewart, Jie Zhao, Nils Lehmann,
  Thomas Dujardin, Zhenghang Yuan, Pedram Ghamisi, and Xiao~Xiang Zhu.
\newblock Geolangbind: Unifying earth observation with agglomerative
  vision-language foundation models.
\newblock \emph{arXiv preprint arXiv:2503.06312}, 2025.

\bibitem[Xu et~al.(2023)Xu, Xie, Tan, Huang, Howes, Sharma, Li, Ghosh,
  Zettlemoyer, and Feichtenhofer]{xu2023metaclip}
Hu Xu, Saining Xie, Xiaoqing~Ellen Tan, Po-Yao Huang, Russell Howes, Vasu
  Sharma, Shang-Wen Li, Gargi Ghosh, Luke Zettlemoyer, and Christoph
  Feichtenhofer.
\newblock Demystifying clip data.
\newblock \emph{arXiv preprint arXiv:2309.16671}, 2023.

\bibitem[Yang and Newsam(2010)]{yang2010ucm}
Yi Yang and Shawn Newsam.
\newblock Bag-of-visual-words and spatial extensions for land-use
  classification.
\newblock In \emph{Proceedings of the 18th SIGSPATIAL international conference
  on advances in geographic information systems}, pages 270--279, 2010.

\bibitem[Yeo et~al.(2025)Yeo, Cha, Song, Jin, and Kim]{yeo2025atas}
Juan Yeo, Soonwoo Cha, Jiwoo Song, Hyunbin Jin, and Taesup Kim.
\newblock Atas: Any-to-any self-distillation for enhanced open-vocabulary dense
  prediction.
\newblock In \emph{Proceedings of the IEEE/CVF International Conference on
  Computer Vision}, pages 20390--20400, 2025.

\bibitem[Yuan et~al.(2022{\natexlab{a}})Yuan, Zhang, Fu, Li, Deng, Wang, and
  Sun]{yuan2022rsitmd}
Zhiqiang Yuan, Wenkai Zhang, Kun Fu, Xuan Li, Chubo Deng, Hongqi Wang, and Xian
  Sun.
\newblock Exploring a fine-grained multiscale method for cross-modal remote
  sensing image retrieval.
\newblock \emph{arXiv preprint arXiv:2204.09868}, 2022{\natexlab{a}}.

\bibitem[Yuan et~al.(2022{\natexlab{b}})Yuan, Zhang, Li, Pan, Mao, Chen, Li,
  Wang, and Sun]{yuan2022slm}
Zhiqiang Yuan, Wenkai Zhang, Chongyang Li, Zhaoying Pan, Yongqiang Mao,
  Jialiang Chen, Shouke Li, Hongqi Wang, and Xian Sun.
\newblock Learning to evaluate performance of multi-modal semantic
  localization.
\newblock \emph{arXiv preprint arXiv:2209.06515}, 2022{\natexlab{b}}.

\bibitem[Zhang et~al.(2024{\natexlab{a}})Zhang, Zhang, Dong, Zang, and
  Wang]{zhang2024longclip}
Beichen Zhang, Pan Zhang, Xiaoyi Dong, Yuhang Zang, and Jiaqi Wang.
\newblock Long-clip: Unlocking the long-text capability of clip.
\newblock In \emph{European conference on computer vision}, pages 310--325.
  Springer, 2024{\natexlab{a}}.

\bibitem[Zhang and Deng(2019)]{zhang2019powerplant}
Haopeng Zhang and Qin Deng.
\newblock Deep learning based fossil-fuel power plant monitoring in high
  resolution remote sensing images: A comparative study.
\newblock \emph{Remote Sensing}, 11\penalty0 (9):\penalty0 1117, 2019.

\bibitem[Zhang et~al.(2019)Zhang, Yuan, Feng, and Lu]{zhang2019hrrsd}
Yuanlin Zhang, Yuan Yuan, Yachuang Feng, and Xiaoqiang Lu.
\newblock Hierarchical and robust convolutional neural network for very
  high-resolution remote sensing object detection.
\newblock \emph{IEEE Transactions on Geoscience and Remote Sensing},
  57\penalty0 (8):\penalty0 5535--5548, 2019.

\bibitem[Zhang et~al.(2024{\natexlab{b}})Zhang, Zhao, Guo, and
  Yin]{zhang2024rs5m}
Zilun Zhang, Tiancheng Zhao, Yulong Guo, and Jianwei Yin.
\newblock Rs5m and georsclip: A large scale vision-language dataset and a large
  vision-language model for remote sensing.
\newblock \emph{IEEE Transactions on Geoscience and Remote Sensing},
  2024{\natexlab{b}}.

\bibitem[Zheng et~al.(2024)Zheng, Zhang, Wu, Lu, Ma, Jin, Chen, and
  Shen]{zheng2024dreamlip}
Kecheng Zheng, Yifei Zhang, Wei Wu, Fan Lu, Shuailei Ma, Xin Jin, Wei Chen, and
  Yujun Shen.
\newblock Dreamlip: Language-image pre-training with long captions.
\newblock In \emph{European Conference on Computer Vision}, pages 73--90.
  Springer, 2024.

\bibitem[Zhong et~al.(2022)Zhong, Yang, Zhang, Li, Codella, Li, Zhou, Dai,
  Yuan, Li, et~al.]{zhong2022regionclip}
Yiwu Zhong, Jianwei Yang, Pengchuan Zhang, Chunyuan Li, Noel Codella,
  Liunian~Harold Li, Luowei Zhou, Xiyang Dai, Lu Yuan, Yin Li, et~al.
\newblock Regionclip: Region-based language-image pretraining.
\newblock In \emph{Proceedings of the IEEE/CVF conference on computer vision
  and pattern recognition}, pages 16793--16803, 2022.

\bibitem[Zhou et~al.(2022)Zhou, Loy, and Dai]{zhou2022maskclip}
Chong Zhou, Chen~Change Loy, and Bo Dai.
\newblock Extract free dense labels from clip.
\newblock In \emph{European conference on computer vision}, pages 696--712.
  Springer, 2022.

\bibitem[Zhou et~al.(2018)Zhou, Newsam, Li, and Shao]{zhou2018patternnet}
Weixun Zhou, Shawn Newsam, Congmin Li, and Zhenfeng Shao.
\newblock Patternnet: A benchmark dataset for performance evaluation of remote
  sensing image retrieval.
\newblock \emph{ISPRS journal of photogrammetry and remote sensing},
  145:\penalty0 197--209, 2018.

\bibitem[Zhu et~al.(2025)Zhu, Wang, Chen, Liu, Ye, Gu, Tian, Duan, Su, Shao,
  et~al.]{zhu2025internvl3}
Jinguo Zhu, Weiyun Wang, Zhe Chen, Zhaoyang Liu, Shenglong Ye, Lixin Gu, Hao
  Tian, Yuchen Duan, Weijie Su, Jie Shao, et~al.
\newblock Internvl3: Exploring advanced training and test-time recipes for
  open-source multimodal models.
\newblock \emph{arXiv preprint arXiv:2504.10479}, 2025.

\bibitem[Zhu et~al.(2018)Zhu, Wen, Bian, Ling, and Hu]{zhu2018visdrone}
Pengfei Zhu, Longyin Wen, Xiao Bian, Haibin Ling, and Qinghua Hu.
\newblock Vision meets drones: A challenge.
\newblock \emph{arXiv preprint arXiv:1804.07437}, 2018.

\end{thebibliography}
}

\clearpage
\setcounter{page}{1}
\maketitlesupplementary

\section{Appendix}

\subsection{Dense Representation}
\label{sec:dense_repre}
To obtain the dense feature map, we follow recent training-free OVSS methods to extract patch-level features from the ViT-based vision encoder \cite{zhou2022maskclip,lan2024clearclip}.
Specifically, we refer to the approach in \cite{li2025segearth}, making slightly modifies to the last residual attention block.
Let the input to the last attention block be denoted by $\mathbf{x}$, which consists of a class embedding $\mathbf{x}_0$ and $h \times w$ patch embeddings. The output feature is computed as
\begin{equation}
\mathbf{y} = \mathrm{CustAttn}(\mathbf{x})
\end{equation}
where $\mathrm{CustAttn}$ represents a customized attention block without residual connection \cite{li2025segearth,lan2024clearclip}. It can be specifically expressed as follows:
\begin{equation}
\mathbf{q} = \mathrm{Emb}_q(\mathbf{x}), \quad
\mathbf{k} = \mathrm{Emb}_k(\mathbf{x}), \quad
\mathbf{v} = \mathrm{Emb}_v(\mathbf{x}).
\end{equation}
\begin{equation}
\mathbf{A}_{tt} = 
\mathrm{Softmax}\!\left( \frac{\mathbf{t}\mathbf{t}^\top}{\sqrt{d_h}} \right),
\quad
t \in \{q, k, v\}
\end{equation}
\begin{equation}
\mathbf{A} = \mathbf{A}_{qq} + \mathbf{A}_{kk} + \mathbf{A}_{vv}
\end{equation}
\begin{equation}
\mathbf{y} = \mathrm{Proj}(\mathbf{A} \mathbf{v})
\end{equation}
where the final attention matrix is denoted by $\mathbf{A}$ and the output feature by $\mathbf{y}$. The final dense feature is obtained by discarding the class embedding $\mathbf{y}_0$ and reshaping the remaining features into an $h \times w$ feature map.


\subsection{Open-Vocabulary Semantic Segmentation}
For the OVSS implementation, we compute logits by comparing the visual features described in \cref{sec:dense_repre} with text queries. The text queries are generated by applying prompt templates to category labels and encoding the resulting descriptions with the text encoder. For comparison methods based on the CLIP architecture, such as CLIP, MetaCLIP, TIPS, and other RS variants, we also use the approach in \cref{sec:dense_repre} to obtain the enhanced dense features. For models trained specifically with dense features optimization, such as CLIPSelf, FineCLIP, and COSMOS, we use the dense features extracted by their own methods.

\subsection{Comprehensive Evaluation of Dense Feature Representations}
\subsubsection{Measuring Feature Discriminability}
\label{sec:feature_discri}
To evaluate the model's discriminability of visual features across different categories, we randomly sample instances for each category and compute the Davies-Bouldin Index (DBI), as defined in \cref{eq:dbi}, which measures the ratio of within-class scatter to between-class separation. 
\begin{equation}
\text{DBI} = \frac{1}{K} \sum_{i=1}^{K} \max_{j \ne i} \left( \frac{S_i + S_j}{M_{ij}} \right)
\label{eq:dbi}
\end{equation}
\noindent where $K$ is the number of categories, $S_i$ and $S_j$ denote the average intra-class distances of category $i$ and $j$, respectively, and $M_{ij}$ is the Euclidean distance between the centroids of categories $i$ and $j$. Features with higher discriminability exhibit lower DBI values.

Specifically, given a semantic segmentation dataset, we randomly select $N$ samples for each target category. We first extract dense features from the images using the image encoder. Then, for each category, instance-level features are obtained by applying average pooling over the pixels within the ground-truth mask. Finally, we compute the DBI across different categories using these instance features. 
In our experiments, considering that the number of samples per category varies across datasets, we select only categories with more than $N$ samples and set $N=256$. We compute the DBI on the test sets of the iSAID, LoveDA, OpenEarthMap, Potsdam, and UAVid datasets, and report the mean values.


\subsubsection{Measuring the Alignment Between Region and Language}
To assess the alignment between regional visual features and text independently, without being affected by visual localization capability, we simplify the task to a zero-shot instance classification task for a given instance feature.
Specifically, we use the instance-level visual features extracted as described in \cref{sec:feature_discri}. The corresponding text features are obtained by encoding templates combined with category labels using the text encoder. Finally, we measure region-text alignment via Top-1 accuracy by comparing each regional visual feature with all category text embeddings. The category and sample settings are the same as described in \cref{sec:feature_discri}.


\subsubsection{Measuring Semantic Coherence}
Semantic coherence has been repeatedly shown to be crucial for dense prediction tasks. To quantitatively assess this property, we formulate it as a binary classification task on pixel pairs, where the goal is to determine whether two pixels belong to the same category. Specifically, we randomly sample pixel pairs from each image and use the semantic segmentation ground truth to assign binary label: 1 if the pixels belong to the same category, 0 otherwise. We then extract dense visual features and compute the feature similarity for each sampled pixel pair. Finally, mean Average Precision (mAP) is computed over the similarity scores of pixel pairs and their corresponding ground-truth labels, serving as a measure of semantic coherence. In our experiments, we randomly select 300 images per dataset and 500 pixel pairs per image. We report the mean mAP over the test sets of the iSAID, LoveDA, OpenEarthMap, Potsdam, Vaihingen, and UAVid datasets.

\subsection{Training Details}
We conducted our experiments on 4×NVIDIA A100 (40GB) GPUs. For the three losses used during training, we consistently set the weights of \Lglo, \Lloc, and \Ldis~to 1.0, 1.0, and 0.1, respectively, whenever they were applied. The detailed hyperparameter settings for our analysis experiments are shown in \cref{tab:paramter1}, and those for \mn~training are provided in \cref{tab:paramter2}.

\begin{table}[ht]
\caption{Hyperparameter settings for analysis experiments. RS5M is used for local-global alignment analysis, while \dn~is used for both region-category and local-global alignment analysis.}
\centering
\begin{tabular}{l|cc}
\toprule
Setting & RS5M-2.5M & \dn \\
\midrule
Learning rate     & $1\mathrm{e}{-6}$ & $4\mathrm{e}{-7}$ \\
Epochs                 & 1                  & 10                 \\
Weight decay      & 0.1                & 1                  \\
Warmup steps           & 1000               & 250                \\
Batch size        & 40                 & 40                 \\
Scheduler              & Cosine             & Cosine             \\
Optimizer              & AdamW              & AdamW              \\
\bottomrule
\end{tabular}
\label{tab:paramter1}
\end{table}

\begin{table}[ht]
\caption{Hyperparameter settings for \mn~training, where stage-1 is trained on the RS5M dataset and stage-2 on the \dn~dataset.}
\centering
\begin{tabular}{l|cc}
\toprule
Setting & RS5M & \dn \\
\midrule
Learning rate     & $1\mathrm{e}{-6}$ & $4\mathrm{e}{-9}$ \\
Epochs                 & 1                  & 10                 \\
Weight decay      & 1                & 1                  \\
Warmup steps           & 1000               & 250                \\
Batch size        & 40                 & 40                 \\
Scheduler              & Cosine             & Cosine             \\
Optimizer              & AdamW              & AdamW              \\
\bottomrule
\end{tabular}
\label{tab:paramter2}
\end{table}

\begin{table*}[ht]
\setlength{\tabcolsep}{4mm} 
\caption{Statistics of the data sources used in our proposed \dn, including LAE-COD \cite{pan2025LAE} and LAE-FOD \cite{pan2025LAE}, as well as Det-10 \cite{liu2024remoteclip}. The ``Category num” for FAIR1M includes 5 super-classes and 37 sub-classes.}
\centering
\begin{tabular}{llcccc}
\toprule
Source & Dataset & Image num & Avg. image size & Object num & Category num \\
\midrule
\multirow{4}{*}{LAE-COD}
    & AID \cite{xia2017aid}  & 5,286 & 600$\times$600 & 34214 & 1,380 \\
    & NWPU-RESISC45 \cite{cheng2017RESISC45} & 28,906 & 256$\times$256 & 82,343 & 1,595 \\
    & SLM \cite{yuan2022slm} & 152 & 1024$\times$1024 & 1,081 & 106 \\
    & EMS \cite{pan2025LAE}  & 2,566 & 1024$\times$1024 & 39,013 & 1,521 \\
\midrule
\multirow{5}{*}{LAE-FOD}
    & DOTAv2 \cite{xia2018dota} & 14,059 & 1024$\times$1024 & 213,728 & 18 \\
    & FAIR1M \cite{sun2022fair1m} & 47,325 & 600$\times$600 & 328,073 & 5(37) \\
    & NWPU-VHR-10 \cite{cheng2014NWPU-VHR} & 395 & 1002$\times$638 & 1,128 & 10 \\
    & Xview \cite{lam2018xview} & 12,318 & 1024$\times$1024 & 326,834 & 53 \\
    & Power Plant \cite{zhang2019powerplant} & 581 & 574$\times$570 & 891 & 4 \\
\midrule
\multirow{7}{*}{Det-10}
    & AUAIR \cite{vujasinovic2020auair} & 32,820 & 1920$\times$1080 & 131,977 & 8 \\
    & CARPK \cite{hsieh2017carpk} & 1,448 & 1280$\times$720 & 89,774 & 1 \\
    & DIOR \cite{li2020DIOR} & 23,463 & 800$\times$800 & 192,465 & 20 \\
    & HRRSD \cite{zhang2019hrrsd} & 21,761 & 992$\times$965 & 57,137 & 13 \\
    & HRSC \cite{liu2017HRSC}& 1,055 & 1112$\times$802 & 2,976 & 1 \\
    & RSOD \cite{long2017rsod} & 936 & 1059$\times$902 & 7,385 & 4 \\
    & Visdrone \cite{zhu2018visdrone} & 6,448 & 1514$\times$992 & 158,821 & 10 \\
\midrule
\textbf{Total} & & \textbf{199,519} & & \textbf{1,667,840} & \textbf{3,147} \\
\bottomrule
\end{tabular}
\label{tab:dataset_statistics}
\end{table*}

\subsection{Details of \dn}

\subsubsection{Data Sources}
Our \dn~dataset is curated from 16 object detection datasets, sourced from LAE-COD \cite{pan2025LAE}, LAE-FOD \cite{pan2025LAE}, and Det-10 \cite{liu2024remoteclip}. The detailed dataset composition is shown in \cref{tab:dataset_statistics}. Specifically, four datasets from LAE-COD contain automatically generated object annotations using large vision or multimodal models, while the remaining datasets are derived from publicly available object detection benchmarks. For more details, please refer to \cite{pan2025LAE, liu2024remoteclip}.

\subsubsection{Caption Generation}
We perform recaptioning using InternVL3 \cite{zhu2025internvl3} to generate global descriptions of images by incorporating both the image itself and its associated object information. We produce two versions of textual descriptions with different lengths: a short version for the original CLIP  \cite{radford2021clip} (supporting 77 tokens) and a long version for LongCLIP \cite{zhang2024longclip} (supporting 248 tokens), aiming to satisfy the demand for long-text inputs in VLFMs. The prompts used to generate the short and long captions are provided in \cref{tab:prompt-short} and \cref{tab:prompt-long}, respectively. Some examples from the \dn~are shown in \cref{fig:dataset}.

\subsubsection{Quality Assessment}
We randomly selected 200 samples to manually evaluate hallucinations in captions generated by the MLLM.
We use a strict evaluation criteria, considering caption that contain any inaccurate or unnatural descriptions as containing hallucinations.
Under these criteria, brief captions can achieve 88\% accuracy, while long captions achieve 80\% accuracy. 
We found that the captions are generally accurate, with high correctness in describing object categories and scene content. Even when containing hallucinations, they have minimal impact on alignment performance.
The most common hallucinations occur in counting, with occasional occurrences in vague spatial descriptions and references to prompt-specific information (\eg, explicitly mentioning the term “bounding box”).
We also observed that in most cases, brief captions are sufficient to describe the scene in detail.

\subsubsection{Ablation Study}
Since our data source includes Det-10 \cite{liu2024remoteclip}, which already provides existing captions, we conduct an ablation study comparing the re-captioned versions with the original captions. Specifically, we fine-tune models on top of CLIP and evaluate zero-shot classification accuracy, as shown in \cref{tab:recaption_original}. The results demonstrate significant improvements in model performance, underscoring the critical role of our high-quality text annotations.

\begin{table*}[h]
\setlength{\tabcolsep}{1mm} 
\caption{Zero-shot classification accuracy comparison of models fine-tuned on CLIP using recaptioned versus original captions}
\centering
\begin{tabular}{lcccccccccccccccc}
\toprule
\multirow{2}{*}{Caption} & \multicolumn{2}{c}{SkyScript} & \multicolumn{2}{c}{AID} & \multicolumn{2}{c}{EuroSAT} & \multicolumn{2}{c}{MillionAID} & \multicolumn{2}{c}{NWPU} & \multicolumn{2}{c}{PatternNet} & \multicolumn{2}{c}{RSICB} & \multicolumn{2}{c}{Mean} \\
 & top1 & top5 & top1 & top5 & top1 & top5 & top1 & top5 & top1 & top5 & top1 & top5 & top1 & top5 & top1 & top5 \\
\midrule
Original  & 32.86 & 63.90 & 59.40 & 89.90 & 37.15 & 77.41 & 48.17 & 74.39 & 54.72 & 82.66 & 52.85 & 79.03 & 28.39 & 70.33 & 44.79 & 76.80 \\
Recaption & 47.88 & 77.95 & 70.05 & 94.75 & 41.85 & 85.96 & 58.99 & 87.86 & 65.31 & 92.92 & 68.69 & 90.73 & 44.38 & 81.89 & 56.74 & 87.44 \\
\bottomrule
\end{tabular}
\label{tab:recaption_original}
\end{table*}

\begin{table}[h]
\caption{Performance comparison of different text lengths under different loss settings. Only \Lglo~is affected by text length.}
\centering
\setlength{\tabcolsep}{1.3mm}
\begin{tabular}{l|cc|cc|cc}
\toprule
\multirow{2}{*}{Text length} & \multirow{2}{*}{\Lglo} & \multirow{2}{*}{\Lloc} & \multicolumn{2}{c|}{ViT-B/16} & \multicolumn{2}{c}{ViT-B/32} \\
 & & & ZSC & OVSS & ZSC & OVSS \\
\midrule
\multirow{2}{*}{Short} & \cmark &        & 56.92 & 35.23 & 54.04 & 30.64 \\
 & \cmark & \cmark & 59.17 & 35.67 & 58.34 & 30.56 \\ \midrule
\multirow{2}{*}{Long}  & \cmark &        & 56.88 & 35.25 & 52.38 & 30.69 \\
  & \cmark & \cmark & 59.24 & 35.66 & 58.14 & 30.44 \\
\bottomrule
\end{tabular}
\label{tab:text-granularity}
\end{table}


\subsection{Effects of Caption Length}

Previous studies have emphasized the importance of using lengthy and detailed textual descriptions for fine-grained vision-language alignment \cite{choi2025goal,asokan2025finelip,xie2025FG-CLIP}. 
To investigate this factor, we compare the effects of global image-caption contrastive learning using short versus long captions from the \dn~dataset.
For long-text encoding, we follow the implementation of Long-CLIP \cite{zhang2024longclip}: the original CLIP positional embeddings for the first 20 tokens are preserved, while the embeddings for positions beyond 20 are extended using a 4× interpolation strategy. This extends the maximum supported input length from 77 to 248 tokens.

\begin{figure*}[t]
  \centering
  \includegraphics[width=0.95\linewidth]{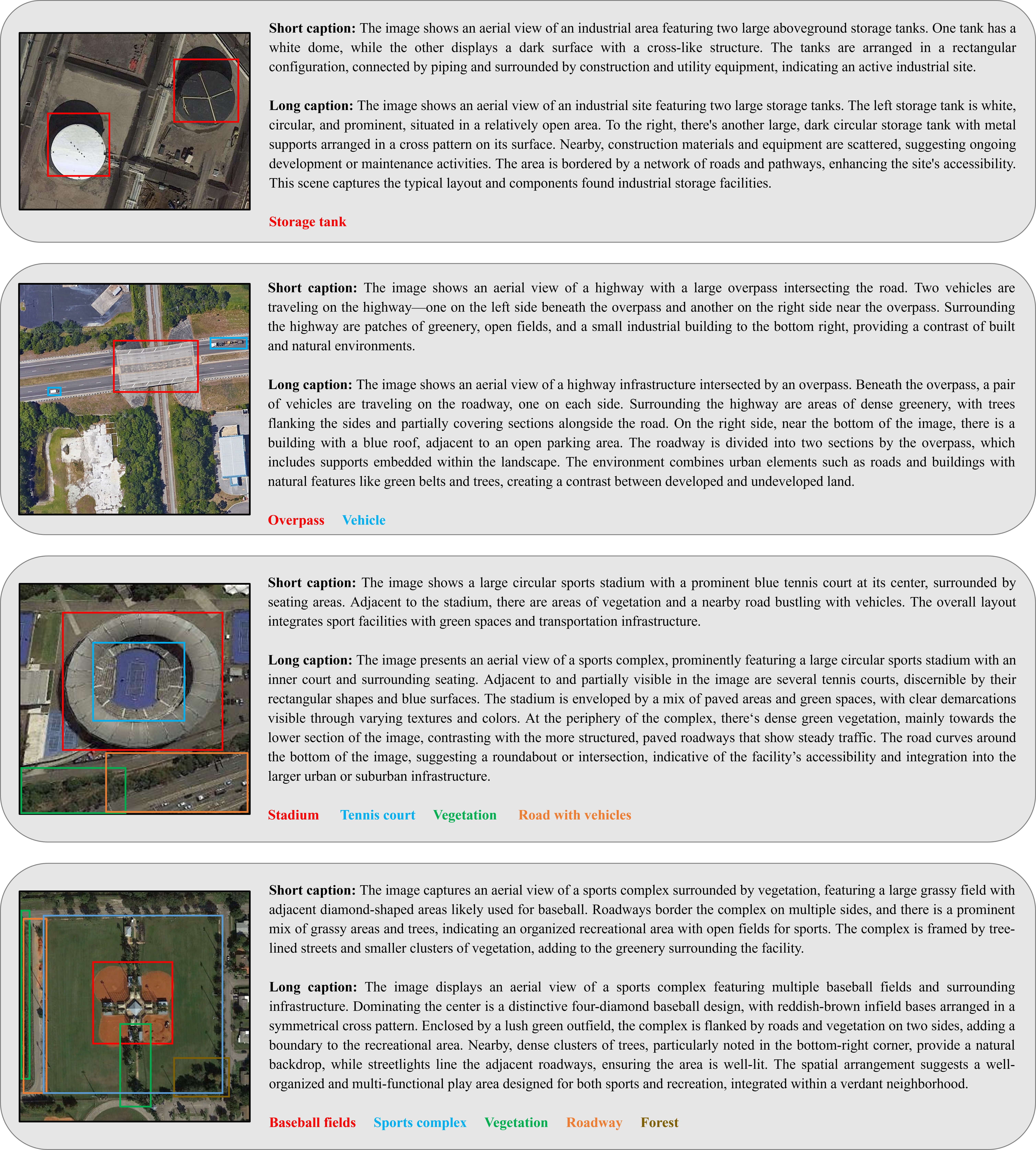}
  \caption{Four examples from our proposed \dn~dataset.}
  \label{fig:dataset}
\end{figure*}

We evaluate the effect of caption length under two settings: using only \Lglo, and using both \Lglo~and \Lloc. Note that caption length only affects \Lglo, as it is trained with global-level image-caption contrastive learning. As shown in \cref{tab:text-granularity}, the model trained with long captions shows negligible difference compared to the one trained with short captions.
We attribute this to the fact that our short captions are generated via MLLM-based recaptioning, rather than being simplistic web-crawled descriptions. 
Despite the 77-token limit, these short captions sufficiently capture the rich semantics of RS images in \dn~dataset, as shown in \cref{fig:dataset}, making the benefits of longer input texts less pronounced.

Nonetheless, we release both the original 77-token-length model and the Long-CLIP variants of our \mn-s2 to accommodate different downstream application needs. In addition, designing methods that specifically leverage long captions for fine-grained alignment \cite{zheng2024dreamlip,xiao2025flair} could better exploit the potential of the data, which we encourage as a direction for future research.

\begin{figure*}[t]
  \centering
  \includegraphics[width=0.95\linewidth]{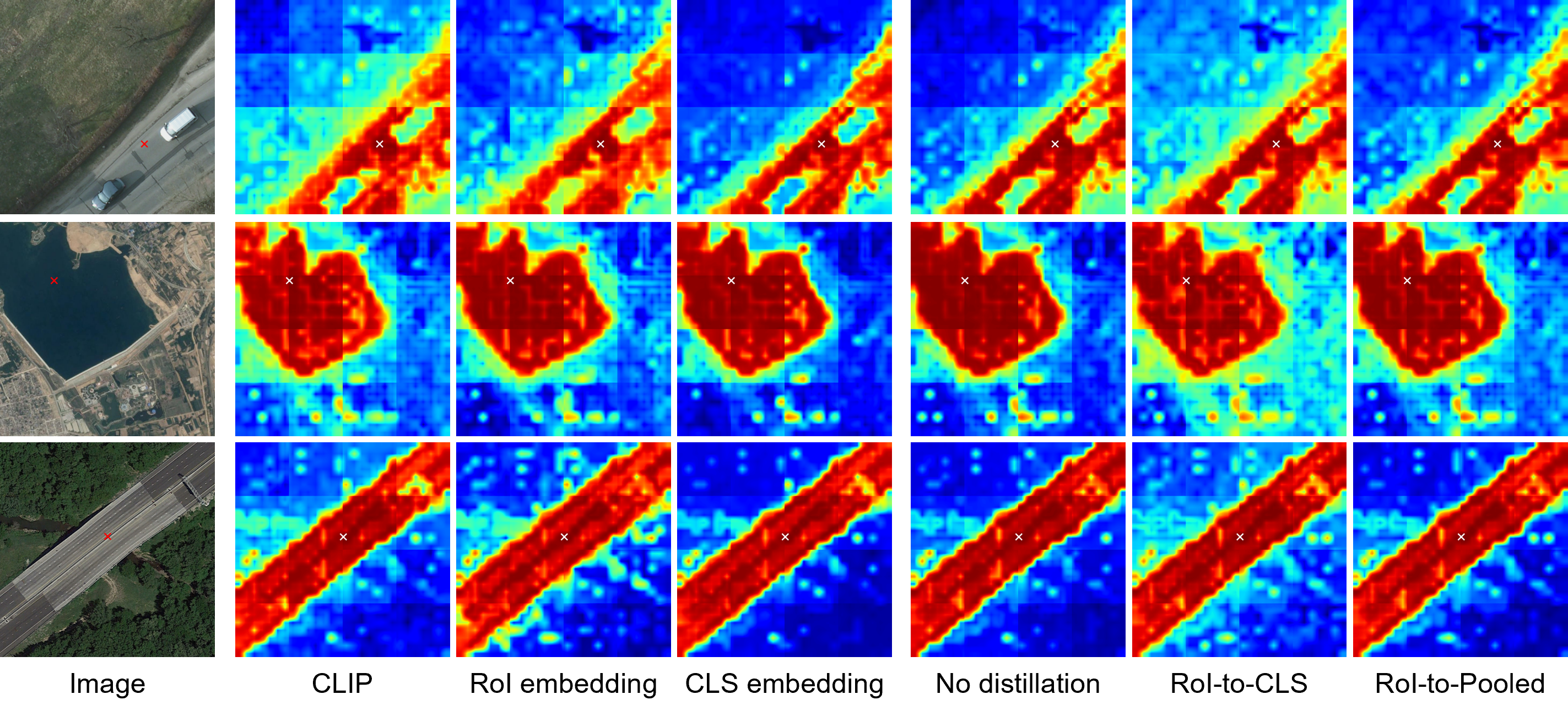}
  \caption{Cosine similarity between the anchor pixel (noted in ×) and all other pixels. Columns 2–4 correspond to the baseline (OpenAI CLIP), ROI embedding-based region-category alignment, and CLS token-based region-category alignment, respectively. ROI-based training tends to disrupt pixel-level semantic coherence, while CLS token-based training enhances it. Columns 5–7 correspond to the baseline (without self-distillation), RoI-to-CLS self-distillation, and RoI-to-Pooled self-distillation, respectively. The CLS token-based approach compromises semantic coherence, whereas RoI-to-Pooled effectively preserves it.}
  \label{fig:pixel_sim}
\end{figure*}

\begin{figure*}[t]
  \centering
  \includegraphics[width=0.95\linewidth]{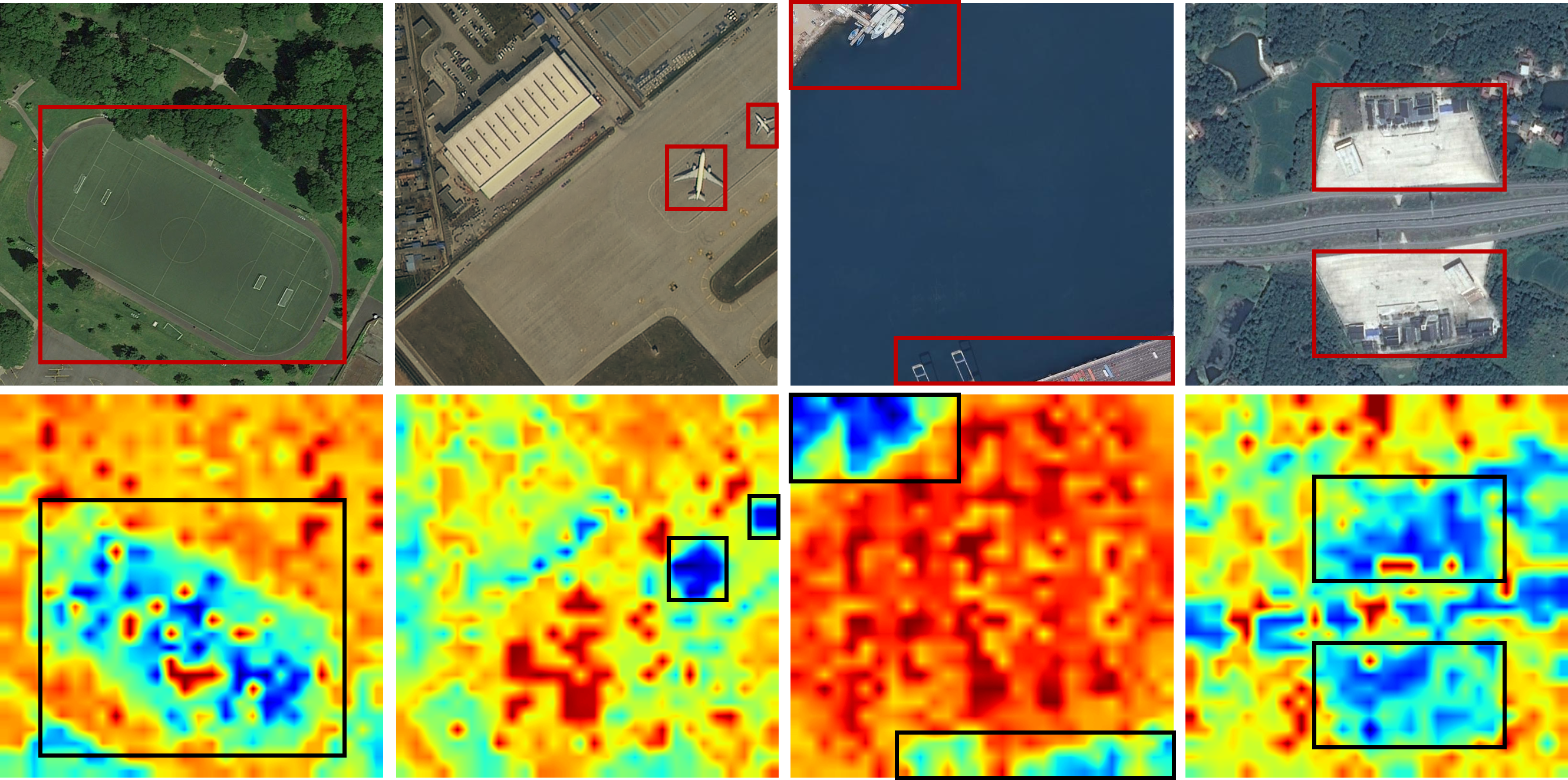}
  \caption{Cosine similarities between the CLS token embedding and all pixels, generated by OpenAI CLIP, show that the CLS token is more similar to the dominant land-cover classes in the image while largely ignoring the relatively minor objects (as indicated by the rectangular annotations). This highlights the risk of losing land-cover recognition in CLS-to-patch self-distillation.}
  \label{fig:c2p_sim}
\end{figure*}

\begin{table*}[h]
    \caption{Prompt for generating short captions based on RS images and corresponding object information.}
    \centering
    \begin{tcolorbox}[width=\textwidth, boxsep=2pt, left=4pt, right=4pt, top=4pt, bottom=4pt]
        \begin{tabular}{p{\dimexpr\textwidth-12pt}}
        
        You are a remote sensing expert specialized in image interpretation and caption generation. You are provided with a remote sensing image and auxiliary data: a list of object bounding boxes in normalized (x1, y1, x2, y2) format (values between 0 and 1) and their corresponding category labels. Your task is to generate a concise and accurate caption that describes the image content, integrating both visual and object-level information. Follow these principles: (1) Generate a brief caption in two or three sentences that describes the image. (2) Focus on the types of objects present, their spatial distribution, and the relationships between them. (3) Do not include any metadata, annotations, or task instructions in the output — only output a natural-language caption.

        \textcolor{cyan!80}{\{\textit{Image}\}}

        \textcolor{cyan!80}{
          \{\textit{Object infos: \;
          Category1: Bbox1; \; Category2: Bbox2,  \ldots
          }\}}
        \end{tabular}
    \end{tcolorbox}
    \label{tab:prompt-short}
\end{table*}

\begin{table*}[t]
    \caption{Prompt for generating long captions based on RS images and corresponding object information.}
    \centering
    \begin{tcolorbox}[width=\textwidth, boxsep=2pt, left=4pt, right=4pt, top=4pt, bottom=4pt]
        \begin{tabular}{p{\dimexpr\textwidth-12pt}}
        
        You are a remote sensing expert specialized in image interpretation and caption generation. You are provided with a remote sensing image and auxiliary data: a list of object bounding boxes in normalized (x1, y1, x2, y2) format (values between 0 and 1) and their corresponding category labels. Your task is to generate a detailed, accurate, and fluent caption that describes the image content, integrating both visual and object-level information. Follow these principles: (1) Detailed describe the image starting with a brief summary of the overall scene or environment. (2) If the number of objects is small, describe their attributes, approximate locations, and inter-object relationships. (3) If the number of objects is large, describe the object distribution, density, and spatial patterns. (4) Do not include any metadata, annotations, or task instructions in the output — only output a natural-language caption.

        \textcolor{cyan!80}{\{\textit{Image}\}}

        \textcolor{cyan!80}{
          \{\textit{Object infos: \;
          Category1: Bbox1; \; Category2: Bbox2,  \ldots
          }\}}
        \end{tabular}
    \end{tcolorbox}
    \label{tab:prompt-long}
\end{table*}

\end{document}